\documentclass{article}

\usepackage{microtype}
\usepackage{graphicx}
\usepackage{amsmath}
\usepackage{amssymb}
\usepackage{amsthm}
\usepackage{booktabs}
\usepackage{subcaption}
\usepackage{enumitem}
\usepackage{flushend}

\newtheorem{theorem}{Theorem}

\newtheorem{definition}{Definition}

\newtheorem{lemma}{Lemma}
\newtheorem{fact}{Fact}

\usepackage{color}

\newcommand{\iR}{\mathbb{R}}
\DeclareMathOperator*{\argmin}{arg\,min}

\newcommand{\R}{\mathbb{R}}

\usepackage[accepted]{icml2018}

\icmltitlerunning{Modeling Sparse Deviations for Compressed Sensing using Generative Models}

\begin{document}

\twocolumn[
\icmltitle{Modeling Sparse Deviations for Compressed Sensing using Generative Models}




\begin{icmlauthorlist}
\icmlauthor{Manik Dhar}{stan}
\icmlauthor{Aditya Grover}{stan}
\icmlauthor{Stefano Ermon}{stan}
\end{icmlauthorlist}

\icmlaffiliation{stan}{Computer Science Department, Stanford University, CA, USA}
\icmlcorrespondingauthor{Manik Dhar}{dmanik@cs.stanford.edu}
\icmlcorrespondingauthor{Aditya Grover}{adityag@cs.stanford.edu}
\icmlkeywords{Machine Learning, ICML}

\vskip 0.3in
]



\printAffiliationsAndNotice{}  

\begin{abstract}
In compressed sensing,  
a small number of linear measurements can be used to reconstruct an unknown signal. Existing approaches leverage assumptions on the structure of these signals, such as sparsity or the availability of a generative model.
A domain-specific generative model can provide a stronger prior and thus allow for recovery with far fewer measurements.
However, unlike sparsity-based approaches, existing methods based on generative models guarantee exact recovery only over their support, which is typically only a small subset of the space on which the signals are defined.
We propose Sparse-Gen, a framework that allows for \emph{sparse} deviations from the support set, thereby achieving the best of both worlds by using a domain specific prior and allowing reconstruction over the full space of signals. Theoretically, our framework provides a new class of signals that can be acquired using compressed sensing, reducing classic sparse vector recovery to a special case and avoiding the restrictive support due to a generative model prior. Empirically, we observe consistent improvements in reconstruction accuracy over competing approaches, especially in the more practical setting of \textit{transfer} compressed sensing where a generative model for a data-rich, source domain aids sensing on a data-scarce, target domain.
\end{abstract}

\section{Introduction}

In many real-world domains, data acquisition is costly. For instance, magnetic resonance imaging (MRI) requires scan times proportional to the number of measurements, which can be significant for patients~\citep{lustig2008compressed}. Geophysical applications like oil drilling require expensive simulation of seismic waves~\citep{qaisar2013compressive}. Such applications, among many others, can benefit significantly from \textit{compressed sensing} techniques to acquire signals efficiently \cite{candes2005decoding,donoho2006compressed,candes2006robust}. 

In compressed sensing, we wish to acquire an $n$-dimensional signal $x \in \mathbb{R}^n$ using only $m \ll n$ measurements linear in $x$. The measurements could potentially be noisy, but even in the absence of any noise we need to impose additional structure on the signal to guarantee unique recovery. Classical results on compressed sensing impose structure by assuming the underlying signal to be approximately $l$-sparse in some known basis, \textit{i.e.}, the $l$-largest entries dominate the rest. For instance, images and audio signals are typically sparse in the wavelet and Fourier basis respectively~\citep{mallat2008wavelet}.
If the matrix of linear vectors relating the signal and measurements satisfies certain mild conditions, then one can provably recover $x$ with only $m=O(l \log \frac{n}{l})$ measurements using LASSO~\cite{tibshirani1996regression,candes2005decoding,donoho2006compressed,candes2006robust,bickel2009simultaneous}.

Alternatively, structural assumptions on the signals being sensed can be \textit{learned} from data, \textit{e.g.}, using a dataset of typical signals~\cite{baraniuk2010model,peyre2010best,chen2010compressive,yu2011statistical}.
Particularly relevant to this work, \citet{bora2017compressed} proposed an approach where structure is provided by a deep generative model learned from data. 
Specifically, the underlying signal $x$ being sensed is assumed to be close to the range of a deterministic function expressed by a pretrained, latent variable model $G: \mathbb{R}^k \rightarrow \mathbb{R}^n $ such that $x \approx G(z)$ where $z \in \mathbb{R}^k$ denote the latent variables. 
Consequently, the signal $x$ is recovered by optimizing for a latent vector $z$ that minimizes the $\ell_2$ distance between the measurements corresponding to $G(z)$ and the actual ones.
Even though the objective being optimized in this case is non-convex, empirical results suggest that the reconstruction error decreases much faster than LASSO-based recovery as we increase the number of measurements.

A limitation of the above approach is that the recovered signal is constrained to be in the range of the generator function $G$. 
Hence, if the true signal being sensed is not in the range of $G$, the algorithm cannot drive the reconstruction error to zero even when $m \geq n$ (even if we ignore error due to measurement noise and non-convex optimization). This is also observed empirically, as the reconstruction error of generative model-based recovery saturates as we keep increasing the number of measurements $m$. On the other hand, LASSO-based recovery continues to shrink the error with increasing number of measurements, eventually outperforming the generative model-based recovery.

To overcome this limitation, we propose a framework that allows recovery of signals with \emph{sparse deviations} from the set defined by the range of the generator function. The recovered signals have the general form of $G(\hat{z})+\hat{\nu}$, where  $\hat{\nu} \in \mathbb{R}^n$ is a sparse vector. This allows the recovery algorithm to consider signals away from the range of the generator function. Similar to LASSO, we relax the hardness in optimizing for sparse vectors by minimizing the $\ell_1$ norm of the deviations. 
Unlike LASSO-based recovery, we can exploit the rich structure imposed by a (deep) generative model (at the expense of solving a hard optimization problem if $G$ is non-convex). In fact, we show that LASSO-based recovery is a special case of our framework if the generator function $G$ maps all $z$ to the origin. Unlike generative model-based recovery, the signals recovered by our algorithm are not constrained to be in the range of the generator function.

Our proposed algorithm, referred to as Sparse-Gen, has desirable theoretical properties and empirical performance.
Theoretically, we derive upper bounds on the reconstruction error for an optimal decoder with respect to the proposed model 
and show that this error vanishes with $m=n$ measurements.
We confirm our theory empirically, wherein we find that recovery using Sparse-Gen with variational autoencoders~\cite{kingma-iclr2014}  as the underlying generative model outperforms both LASSO-based and generative model-based recovery in terms of the reconstruction errors for the same number of measurements for MNIST and Omniglot datasets. Additionally, we observe significant improvements in the more practical and novel task of \textit{transfer} compressed sensing 
where a generative model on a data-rich, source domain provides a prior for sensing a data-scarce, target domain.

\section{Preliminaries}

In this section, we review the necessary background and prior work in modeling domain specific structure in compressed sensing. 
We are interested in solving the following system of equations,
\begin{align}\label{eq:cs}
y &= A x
\end{align}
where $x \in \mathbb{R}^n$ is the signal of interest being sensed through measurements $y \in \mathbb{R}^m$, and $A \in \mathbb{R}^{m \times n}$ is a measurement matrix. For efficient acquisition of signals, we will design measurement matrices such that $m \ll n$. However, the system is under-determined whenever $\textrm{rank}(A) < n$. Hence, unique recovery requires additional assumptions on $x$. We now discuss two ways to model the structure of $x$.

\textbf{Sparsity.} Sparsity in a well-chosen basis is natural in many domains. For instance, natural images are sparse in the wavelet basis whereas audio signals exhibit sparsity in the Fourier basis~\citep{mallat2008wavelet}. Hence, it is natural to assume the domain of signals $x$ we are interested in recovering is
\begin{align}\label{eq:model_lasso}
S_l(0) = \{x: \Vert x - 0\Vert_0\le l\}.
\end{align}
This is the set of $l$-sparse vectors with the $\ell_0$ distance measured from the origin. Such assumptions dominate the prior literature in compressed sensing and can be further relaxed to recover \textit{approximately} sparse signals~\cite{candes2005decoding,donoho2006compressed,candes2006robust}. 

\textbf{Latent variable generative models.} A latent variable model specifies a joint distribution $P_\theta(x, z)$ over the observed data $x$ (\textit{e.g.}, images) and a set of latent variables $z \in \mathbb{R}^k$ (\textit{e.g.}, features). Given a training set of signals $\{x_1, \cdots, x_M\}$, we can learn the parameters $\theta$ of such a model, \textit{e.g.}, via maximum likelihood. When $P_\theta(x, z)$  is parameterized using deep neural networks, such \textit{generative} models can effectively model complex, high-dimensional signal distributions for modalities such as images and audio~\citep{kingma-iclr2014,goodfellow2014generative}.

Given a pretrained latent variable generative model with parameters $\theta$, we can associate a \emph{generative model function} $G:\mathbb{R}^k \rightarrow \mathbb{R}^n$ mapping a latent vector $z$ to the mean of the conditional distribution $P_\theta(x\vert z)$. 
Thereafter, the space of signals that can be recovered with such a model is given by the range of the generator function,
\begin{align}\label{eq:model_gen}
S_G= \{G(z) : z \in R^k\}.
\end{align}
Note that the set is defined with respect to the latent vectors $z$, and we omit the dependence of $G$ on the parameters $\theta$ (which are fixed for a pretrained model) for brevity.
\subsection{Recovery algorithms}
Signal recovery in compressed sensing algorithm typically involves solving an optimization problem consistent with the modeling assumptions on the domain of the signals being sensed. 

\textbf{Sparse vector recovery using LASSO.} Under the assumptions of sparsity, the signal $x$ can be recovered by solving an $\ell_0$ minimization problem~\cite{candes2005decoding,donoho2006compressed,candes2006robust}.
\begin{align}
\min_x \Vert x \Vert_0 \nonumber\\
\textrm{s.t. } Ax=y.
\end{align}
The objective above is however NP-hard to optimize, and hence, it is standard to consider a convex relaxation,
\begin{align}\label{eq:l1}
\min_x \Vert x \Vert_1 \nonumber\\
\textrm{s.t. } Ax=y.
\end{align}
In practice, it is common to solve the Lagrangian of the above problem. We refer to this method as LASSO-based recovery due to similarities of the objective in Eq.~\eqref{eq:l1} to the LASSO regularization used broadly in machine learning~\cite{tibshirani1996regression}. LASSO-based recovery is the predominant technique for recovering sparse signals since it involves solving a tractable convex optimization problem.

In order to guarantee unique recovery to the underdetermined system in Eq.~\eqref{eq:cs}, the measurement matrix $A$ is designed to satisfy the Restricted Isometry Property (RIP) or the Restricted Eigenvalue Condition (REC) for $l$-sparse matrices with high probability~\cite{candes2005decoding,bickel2009simultaneous}.  We define these conditions below.
\begin{definition}\label{def:rip_def}
Let $S_l(0) \subset \mathbb{R}^n$ be the set of $l$-sparse vectors. For some parameter $\alpha\in (0,1)$, a matrix $A \in \mathbb{R}^{m \times n}$ is said to satisfy RIP$(l,\alpha)$ if $\ \forall \ x \in S_l(0)$,
  $$(1 - \alpha)\|x \|_2\leq \|A x\|_2 \leq (1 + \alpha)\|x \|_2.$$
\end{definition}
\begin{definition}\label{def:rec_def}
Let $S_l(0) \subset \mathbb{R}^n$ be the set of $l$-sparse vectors. For some parameter $\gamma>0$, a matrix $A \in \mathbb{R}^{m \times n}$ is said to satisfy REC$(l,\gamma)$ if $\ \forall \ x \in S_l(0)$,
  $$\|A x\|_2 \geq \gamma \|x \|_2.$$
\end{definition}
Intuitively, RIP implies that $A$ approximately preserves Euclidean norms for sparse vectors and REC implies that sparse vectors are far from the nullspace of $A$. Many classes of matrices satisfy these conditions with high probability, including random Gaussian and Bernoulli matrices where every entry of the matrix is sampled from a standard normal and uniform 
Bernoulli distribution respectively~\citep{baraniuk2008simple}. 

\textbf{Generative model vector recovery using gradient descent.} If the signals being sensed are assumed to lie close to the range $S_G$ of a generative model function $G$ as defined in Eq.~\eqref{eq:model_gen}
, then we can recover the best approximation to the true signal by $\ell_2$-minimization over $z$,
\begin{align}\label{eq:gen}
\min_{z} \Vert AG(z) - y \Vert_2^2.
\end{align}
The function $G$ is typically expressed as a deep neural network which makes the overall objective non-convex, but differentiable almost everywhere w.r.t $z$. In practice, good reconstructions can be recovered by gradient-based optimization methods.
We refer to this method proposed by \citet{bora2017compressed} as \textit{generative model-based recovery}. 

 To guarantee unique recovery, generative model-based recovery makes two key assumptions. First, the generator function $G$ is assumed to be $L$-Lipschitz, \textit{i.e.}, $\forall \ z_1, z_2 \in \mathbb{R}^k$,
\[
    \| G(z_1) - G(z_2)\|_2 \leq L \|z_1 - z_2\|_2.
\]
Secondly, the measurement matrix $A$ is designed to satisfy the Set-Restricted Eigenvalue Condition (S-REC) with high probability~\citep{bora2017compressed}.
\begin{definition}\label{def:srec_def}
  Let $S \subseteq \mathbb{R}^n$.  For some parameters $\gamma > 0$,
  $\delta \geq 0$, a matrix $A \in \mathbb{R}^{m \times n}$ is said to
  satisfy the S-REC$(S, \gamma, \delta)$ if  $\ \forall \ x_1, x_2 \in S$,
  $$\Vert A (x_1 - x_2) \Vert_2\geq \gamma \Vert x_1 - x_2 \Vert_2 - \delta.$$
\end{definition}
S-REC generalizes REC to an arbitrary set of vectors $S$ as opposed to just considering the set of approximately sparse vectors $S_l(0)$ and allowing an additional slack term $\delta$. 
In particular, $S$ is chosen to be the range of the generator function $G$ for generative model-based recovery. 
\section{The Sparse-Gen framework}

\begin{figure*}[t]
\centering
\begin{subfigure}[b]{0.32\textwidth}
\centering
\includegraphics[width=0.7\textwidth]{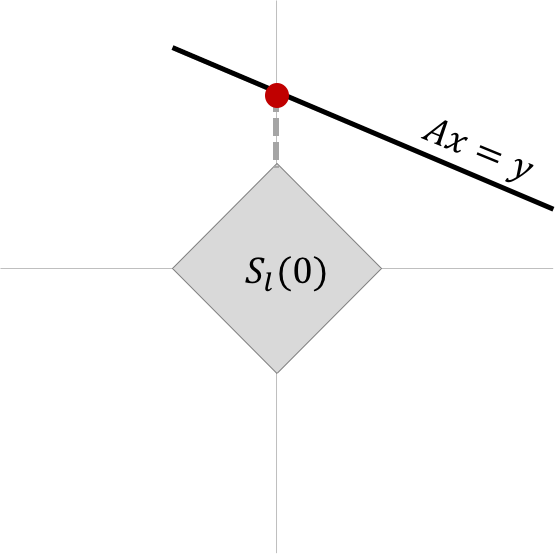}
\caption{LASSO}
\end{subfigure}
\begin{subfigure}[b]{0.32\textwidth}
\centering
\includegraphics[width=0.7\columnwidth]{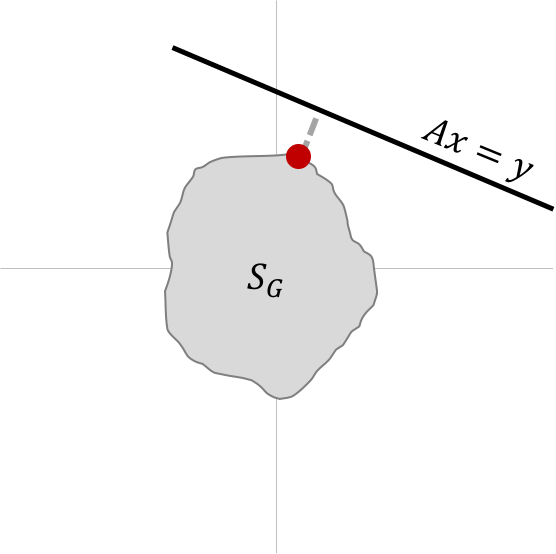}
\caption{Generative model}
\end{subfigure}
\begin{subfigure}[b]{0.32\textwidth}
\centering
\includegraphics[width=0.7\columnwidth]{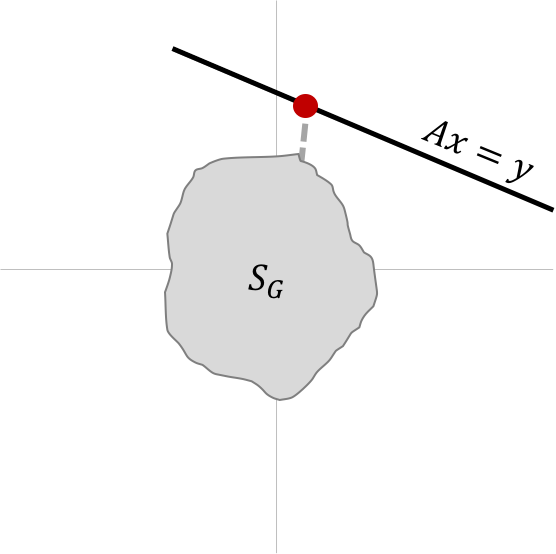}
\caption{Sparse-Gen}
\end{subfigure}
\caption{LASSO vs. Generative model vs. Sparse-Gen recovery. Unlike LASSO, Sparse-Gen imposes a stronger prior on the signals being sensed (\textbf{shaded grey regions}). Unlike generative model, the recovered signals are not constrained to lie on the range of the generator function (\textbf{red points}). Aspects of visualization due to eigenvalue conditions not shown for simplicity.}\label{fig:comparison} 
\end{figure*}

The modeling assumptions based on sparsity and generative modeling discussed in the previous section can be limiting in many cases. On one hand, sparsity assumes a relatively weak prior over the signals being sensed. Empirically, we observe that the recovered signals $x_L$ have large reconstruction error $\Vert x_L - x \Vert_2^2$ especially when the number of measurements $m$ is small. On the other hand, generative models imposes a very strong, but rigid prior which works well when the number of measurements is small. However, the performance of the corresponding recovery methods saturates with increasing measurements since the recovered signal $x_G = G(z_G)$ is constrained to lie in the range of the generator function $G$. If $z_G \in \mathbb{R}^k$ is the optimum value returned by an optimization procedure for Eq.~\eqref{eq:gen}, then the reconstruction error $\Vert x_G - x \Vert_2^2$ is limited by the dimensionality of the latent space and the quality of the generator function.

To sidestep the above limitations, we consider a strictly more expressive class of signals by allowing sparse deviations from the range of a generator function. Formally, the domain of the recovered signals is given by,
\begin{align}\label{eq:model_lassogen}
S_{l,G} = \cup_{z\in \mathrm{Dom}(G)} S_l(G(z))
\end{align}
where $S_l(G(z))$ denotes the set of sparse vectors centered on $G(z)$ and $z$ varies over the domain of $G$ (typically $\mathbb{R}^k$).
We refer to this modeling assumption and the consequent algorithmic framework for recovery as \textit{Sparse-Gen}.

Based on this modeling assumption, we will recover signals of the form $G(z) + \nu$ for some $\nu \in \mathbb{R}^n$ that is preferably sparse. Specifically, we consider the optimization of a hybrid objective,
\begin{align}
\min_{z, \nu}  \Vert \nu \Vert_0 \nonumber\\
\textrm{s.t. } A \left(G(z) + \nu\right) = y.
\end{align}
In the above optimization problem the objective is non-convex and non-differentiable, while the constraint is non-convex (for general $G$), making the above optimization problem hard to solve. To ease the optimization problem, we propose two modifications. First, we relax the $\ell_0$ minimization to an $\ell_1$ minimization similar to LASSO.
\begin{align}
\min_{z, \nu}  \Vert \nu \Vert_1 \nonumber\\
\textrm{s.t. } A \left(G(z) + \nu\right) = y.
\end{align}
Next, we square the non-convex constraint on both sides and consider the Lagrangian of the above problem to get the final unconstrained optimization problem for Sparse-Gen,
\begin{align}\label{eq:lassogen}
\min_{z, \nu} \Vert \nu \Vert_1 + \lambda \Vert A \left(G(z) + \nu\right) - y \Vert_2^2  
\end{align}
where $\lambda$ is the Lagrange multiplier. 

The above optimization problem is non-differentiable w.r.t. $\nu$ and non-convex w.r.t. $z$ (if $G$ is non-convex). In practice, it can be solved in practice using gradient descent (since the non-differentiability is only at a finite number of points) or using sequential convex programming (SCP). SCP is an effective heuristic for non-convex problems where the convex portions of the problem are solved using a standard convex optimization technique~\cite{boyd2004convex}.  In the case of Eq.~\eqref{eq:lassogen}, the  optimization w.r.t. $\nu$ (for fixed $z$) is a convex optimization problem whereas the non-convexity  typically involves differentiable terms (w.r.t. $z$) if $G$ is a deep neural network. 
Empirically, we find excellent recovery by standard first order gradient-based methods
~\cite{duchi2011adaptive,tieleman2012lecture,kingma2014adam}.

Unlike LASSO-based recovery which recovers only sparse signals, Sparse-Gen can impose a stronger domain-specific prior using a generative model. If we fix the generator function to map all $z$ to the origin, we recover LASSO-based recovery as a special case of Sparse-Gen. 
Additionally, Sparse-Gen is not constrained to recover signals over the range of $G$, as in the case of generative model-based recovery. In fact, it can recover signals with sparse deviations from the range of $G$. Note that
the sparse deviations can be defined in a basis different from the canonical basis. In such cases, we consider the following optimization problem,
\begin{align}\label{eq:lassogen_nonstandard}
\min_{z, \nu} \Vert B\nu \Vert_1  + \lambda \Vert A \left(G(z) + \nu\right) - y \Vert_2^2  
\end{align} 
where $B$ is a change of basis matrix that promotes sparsity of the vector $B\nu$. Figure~\ref{fig:comparison} illustrates the differences in modeling assumptions between Sparse-Gen and other frameworks. 
\section{Theoretical Analysis}

The proofs for all results in this section are given in the Appendix. Our analysis and experiments account for measurement noise $\epsilon$ in compressed sensing, \textit{i.e.},
\begin{align}
y &= A x + \epsilon.
\end{align}
Let $\Delta:\iR^m\rightarrow \iR^n$ denote an arbitrary decoding function used to recover the true signal $x$ from the measurements $y \in \iR^m$. Our analysis will upper bound the $\ell_2$-error in recovery incurred by our proposed framework using mixed norm guarantees (in particular, $\ell_2/\ell_1$). To this end, we first state some key definitions. Define the least possible $\ell_1$ error for recovering $x$ under the Sparse-Gen modeling as,
$$\sigma_{S_{l,G}}(x)=\inf_{\hat{x}\in S_{l,G}}\|x-\hat{x}\|_1$$
where the optimal $\hat{x}$ is the closest point to $x$ in the allowed domain $S_{l,G}$. We now state the main lemma guiding the theoretical analysis. 

\begin{lemma}\label{thm:decoder}
Given a function $G:\iR^k\rightarrow \iR^n$ and measurement noise $\epsilon$ with $\|\epsilon\|_2\le \epsilon_{\max}$, let $A$ be any matrix that satisfies S-REC($S_{1.5l,G},(1-\alpha),\delta$) and RIP($2l,\alpha$) for some $\alpha \in (0,1)$, $l>0$. Then, there exists a decoder $\Delta:\iR^m\rightarrow \iR^n$ such that,
$$\|x-\Delta(Ax+\epsilon)\|_2\le (2l)^{-1/2}C_0\sigma_{l,G}(x)+C_1 \epsilon_{\max}+\delta'$$
for all $x\in \iR^n$, where $C_0=2((1+\alpha)(1-\alpha)^{-1}+1),C_1=2(1-\alpha)^{-1}$, and $\delta'=\delta(1-\alpha)^{-1}$.
\end{lemma}

The above lemma shows that there exists a decoder such that the error in recovery can be upper bounded for measurement matrices satisfying S-REC and RIP. Note that Lemma~\ref{thm:decoder} only guarantees the existence of such a decoder and does not prescribe an optimization algorithm for recovery. Apart from the errors due to the bounded  measurement noise $\epsilon_{\max}$ and a scaled slack term appearing in the S-REC condition $\delta'$, the major term in the upper bound corresponds to (up to constants) the minimum possible error incurred by the best possible recovery vector in $S_{l,G}$ given by $\sigma_{l,G}(x)$. Similar terms appear invariably in the compressed sensing literature and are directly related to the modeling assumptions regarding $x$ (for example, Theorem 8.3 in \citet{cohen2009compressed}). 

Our next lemma shows that random Gaussian matrices satisfy the S-REC (over the range of Lipschitz generative model functions) and RIP conditions with high probability for $G$ with bounded domain, both of which together are sufficient conditions for Lemma~\ref{thm:decoder} to hold. 

\begin{lemma}\label{thm:gaussian}
Let $G:B^k(r)\rightarrow \iR^n$ be an $L$-Lipschitz function where $B^k(r) = \lbrace z \ \vert \ z \in \mathbb{R}^k, \| z \|_2 \leq r \rbrace$ is the $\ell_2$-norm ball in $\mathbb{R}^k$. For $\alpha \in (0,1)$, if
$$m=O\left(\frac{1}{\alpha^2}\Big(k\log\left(\frac{Lr}{\delta}\right)+l\log(n/l)\Big)\right) $$
then a random matrix $A \in \mathbb{R}^{m\times n}$ with i.i.d. entries such that $A_{ij} \sim \mathcal{N}\left(0,\frac{1}{m}\right)$ satisfies the S-REC($S_{1.5l,G},1-\alpha,\delta$) and RIP($2l,\alpha$) with $1 - e^{-\Omega(\alpha^2 m)}$ probability.
\end{lemma}

Using Lemma~\ref{thm:decoder} and Lemma~\ref{thm:gaussian}, we can bound the error due to decoding with generative models and random Gaussian measurement matrices in the following result.
\begin{theorem}\label{thm:decoder_gaussian}
Let $G:B^k(r)\rightarrow \iR^n$ be an $L$-Lipschitz function. For any 
$\alpha \in (0,1)$, $l>0$, let $A \in \R^{m\times n}$ be a random Gaussian matrix with $$m=O\left(\frac{1}{\alpha^2}\Big(k\log\left(\frac{Lr}{\delta}\right)+l\log(n/l)\Big)\right) $$ rows of i.i.d. entries scaled such that $A_{i,j} \sim N(0, 1/m)$. 
Let $\Delta$ be the decoder satisfying Lemma~\ref{thm:decoder}. Then, we have with $1 - e^{-\Omega(\alpha^2 m)}$ probability,
$$\|x-\Delta(Ax+\epsilon)\|_2\le (2l)^{-1/2}C_0\sigma_{l,G}(x)+C_1 \epsilon_{\max}+\delta'$$
for all $x\in \R^n, \|\epsilon\|_2\le \epsilon_{\max}$, where $C_0,C_1,\gamma,\delta'$ are constants defined in Lemma~\ref{thm:decoder}.
\end{theorem}

From the above lemma, we see that the number of measurements needed to guarantee upper bounds on the reconstruction error of any signal with high probability depends on two terms. The first term includes dependence on the Lipschitz constant $L$ of the generative model function G. A high Lipschitz constant makes recovery harder 
 (by requiring a larger number of measurements),
 but only contributes logarithmically. The second term, typical of results in sparse vector recovery, shows a logarithmic growth on the dimensionality $n$ of the signals. Ignoring logarithmic dependences and constants, recovery using Sparse-Gen requires about $O(k+l)$ measurements for recovery.
Note that Theorem~\ref{thm:decoder_gaussian} assumes access to an optimization oracle for decoding. In practice, we consider the solutions returned by gradient-based optimization methods to a non-convex objective defined in Eq.~\eqref{eq:lassogen_nonstandard} that are not guaranteed to correspond to the optimal decoding in general. 

Finally, we obtain tighter bounds for the special case when $G$ is expressed using a neural network with only ReLU activations. These bounds do not rely explicitly on the Lipschitz constant $L$ or require the domain of $G$ to be bounded. 

\begin{theorem}\label{thm:decoder_gaussian_relu}
If $G:\iR^k\rightarrow \iR^n$ is a neural network of depth $d$ with only ReLU activations and at most $c$ nodes in each layer, then the guarantees of Theorem~\ref{thm:decoder_gaussian} hold for 
\[
m=O\left(\frac{1}{\alpha^2}\Big((k+l) d \log c+(k+l)\log(n/l)\Big)\right).
\]
\end{theorem}

\begin{figure*}[ht]
\centering
\begin{subfigure}[b]{0.32\textwidth}
\centering
\includegraphics[width=\columnwidth]{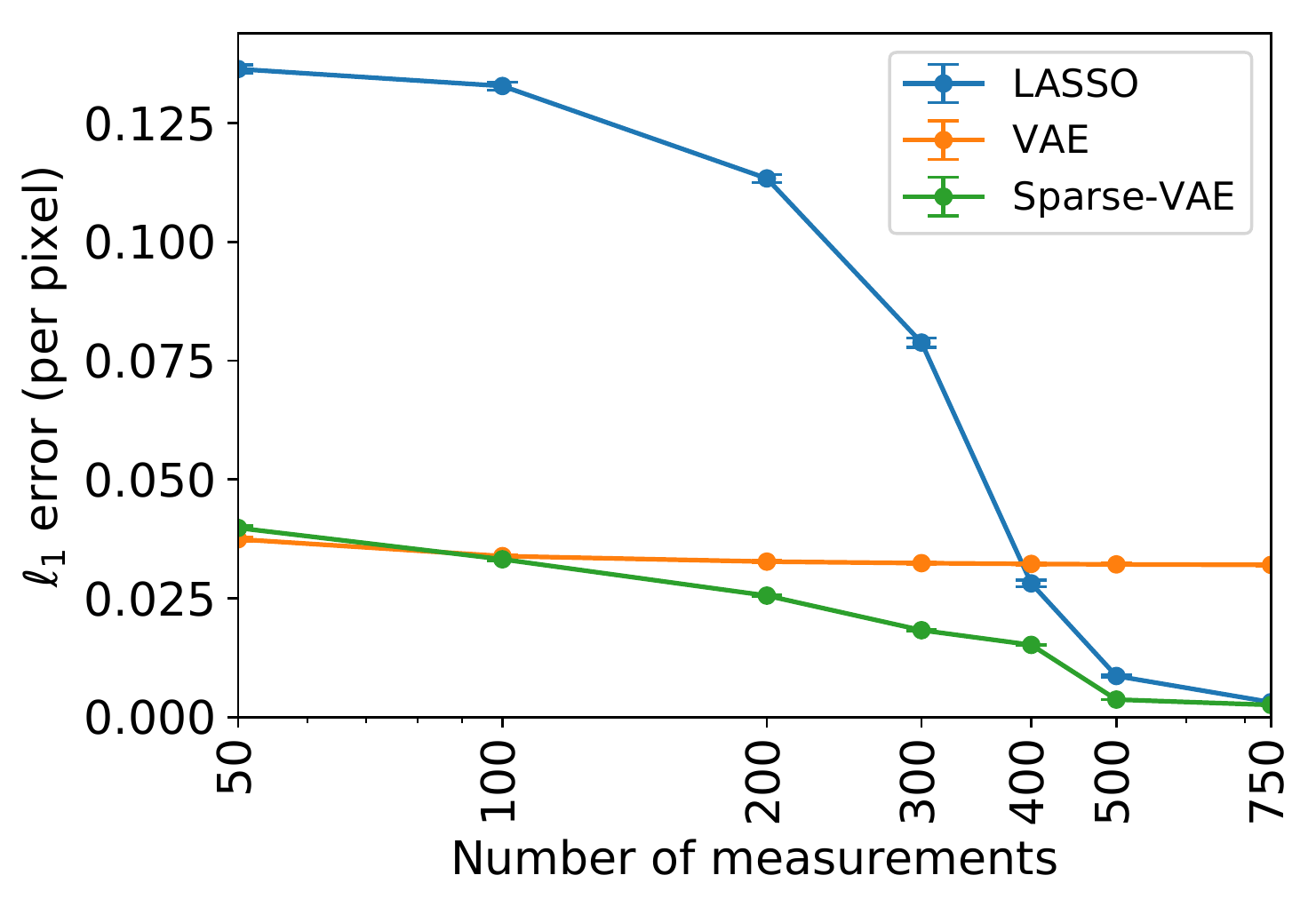}
\caption{MNIST - $\ell_1$}
\end{subfigure}
\begin{subfigure}[b]{0.32\textwidth}
\centering
\includegraphics[width=\columnwidth]{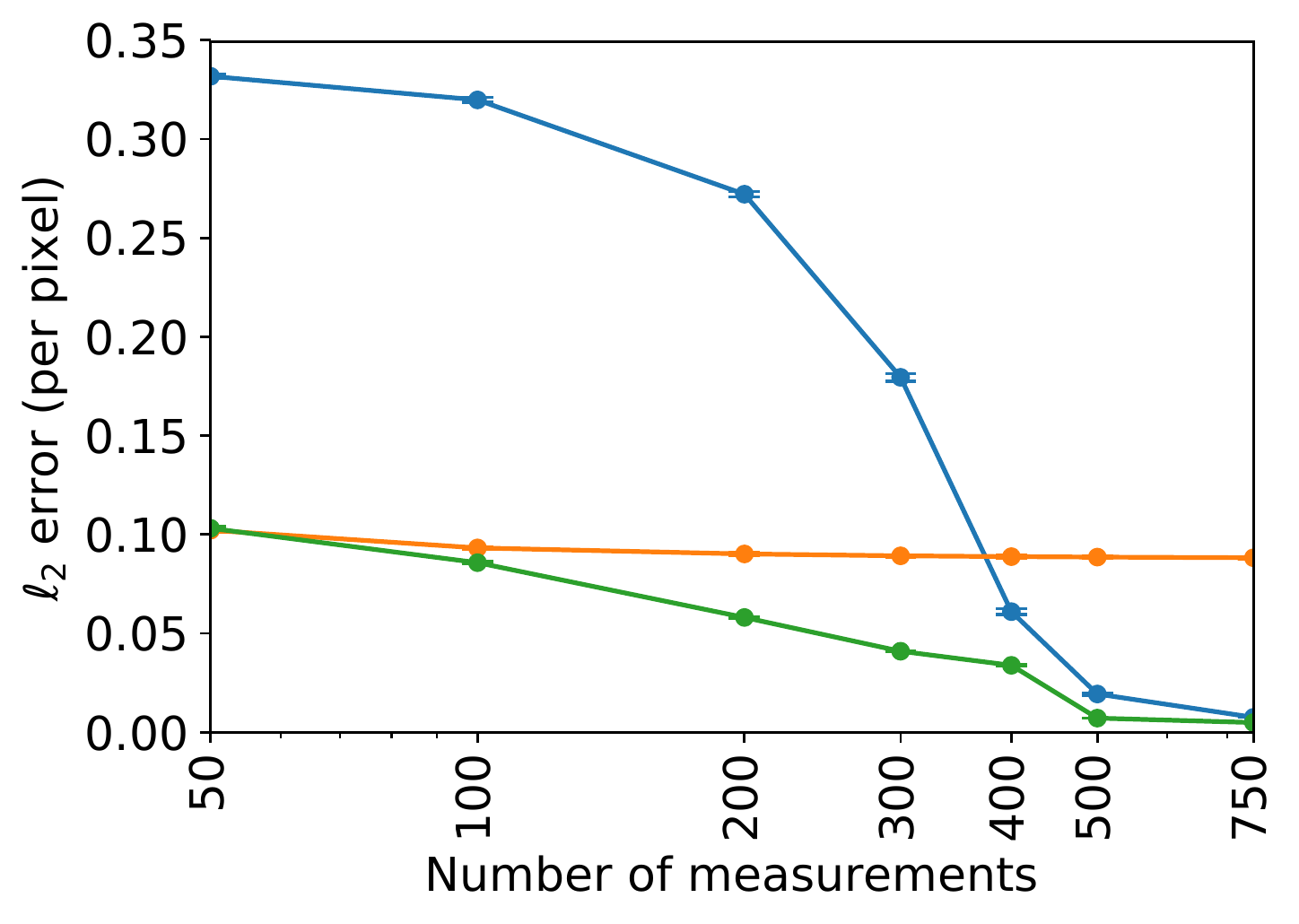}
\caption{MNIST - $\ell_2$}
\end{subfigure}
\begin{subfigure}[b]{0.32\textwidth}
\centering
\includegraphics[width=\columnwidth]{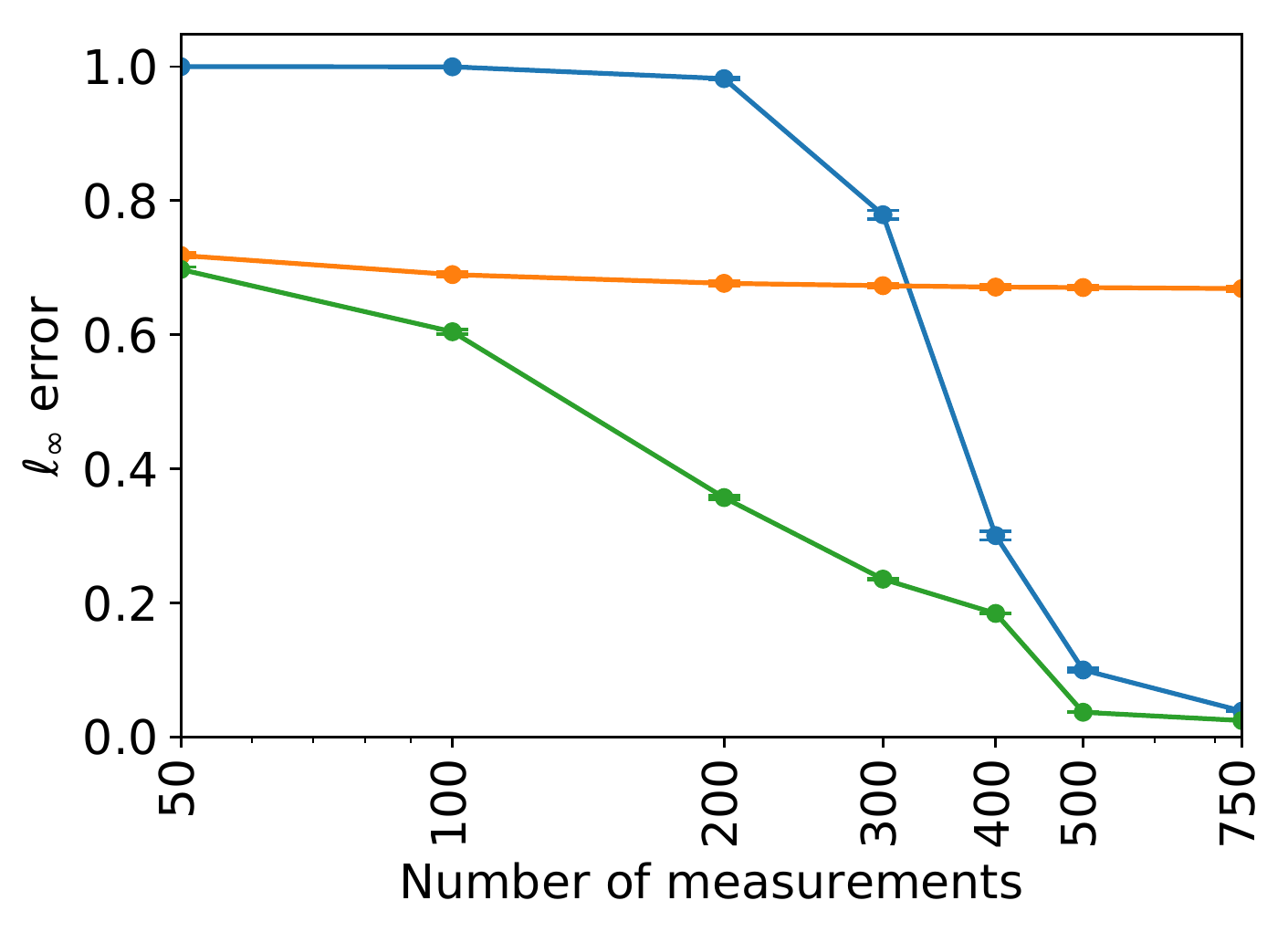}
\caption{MNIST - $\ell_\infty$}
\end{subfigure}
\centering
\begin{subfigure}[b]{0.32\textwidth}
\centering
\includegraphics[width=\columnwidth]{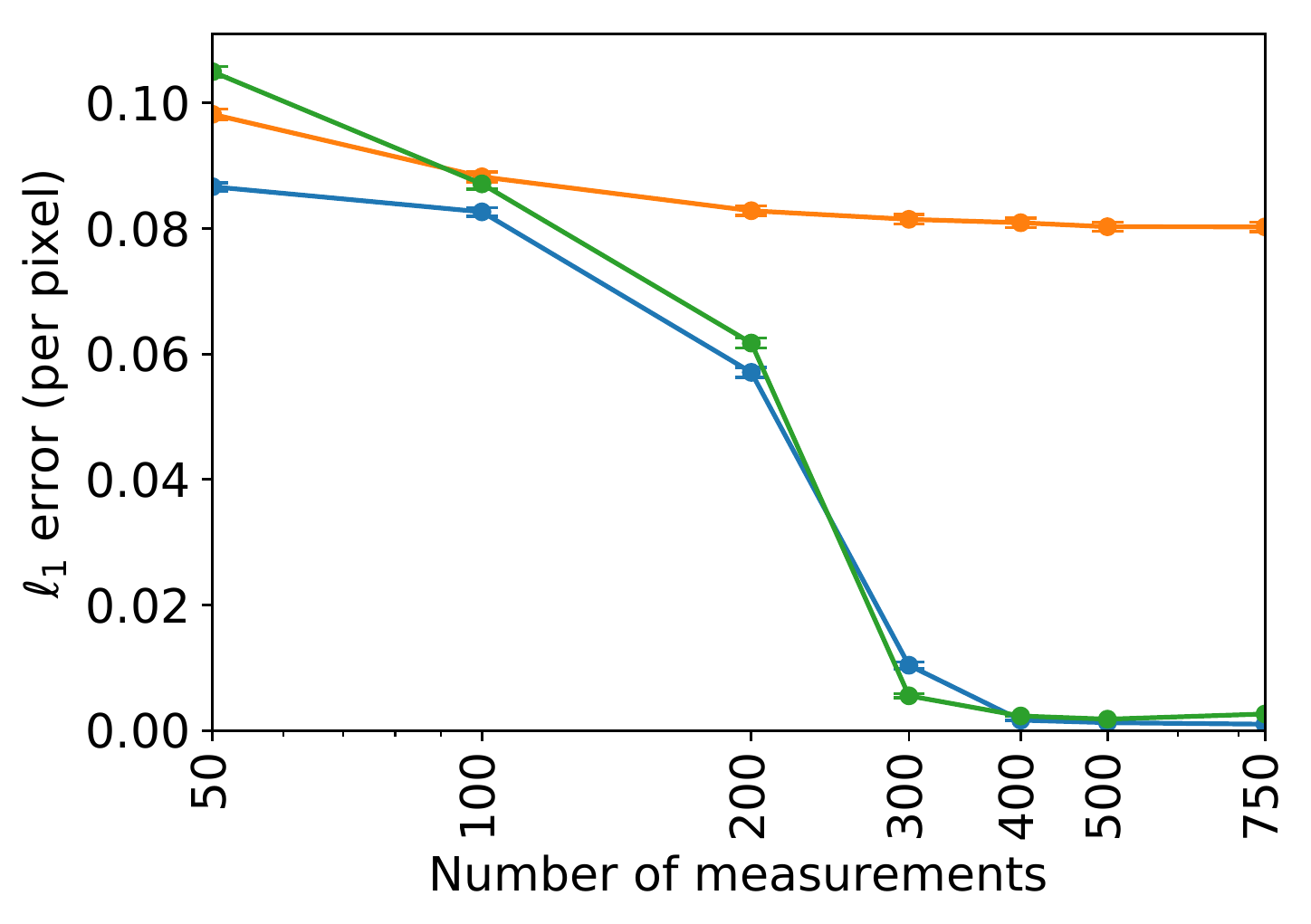}
\caption{Omniglot - $\ell_1$}
\end{subfigure}
\begin{subfigure}[b]{0.32\textwidth}
\centering
\includegraphics[width=\columnwidth]{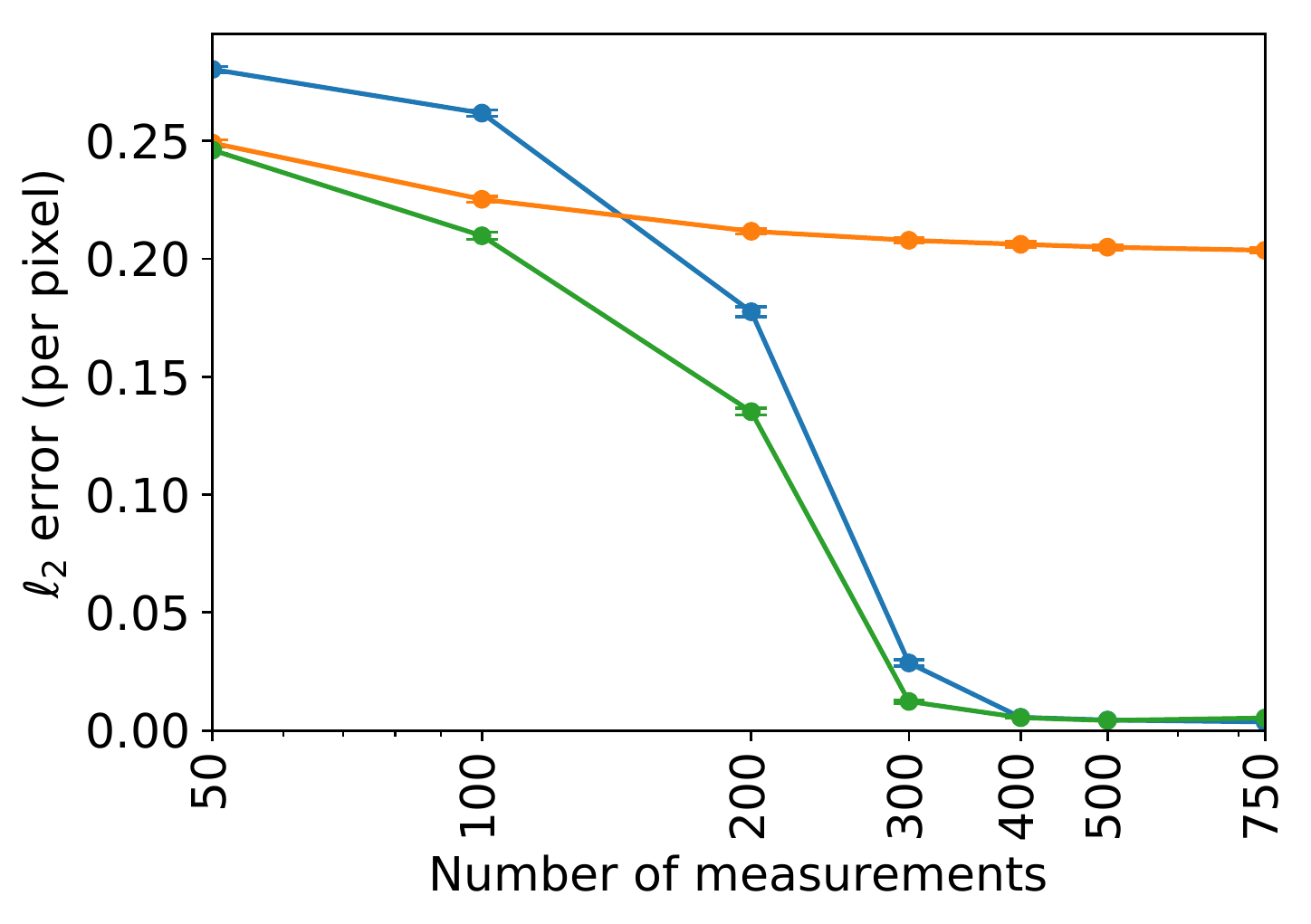}
\caption{Omniglot - $\ell_2$}
\end{subfigure}
\begin{subfigure}[b]{0.32\textwidth}
\centering
\includegraphics[width=\columnwidth]{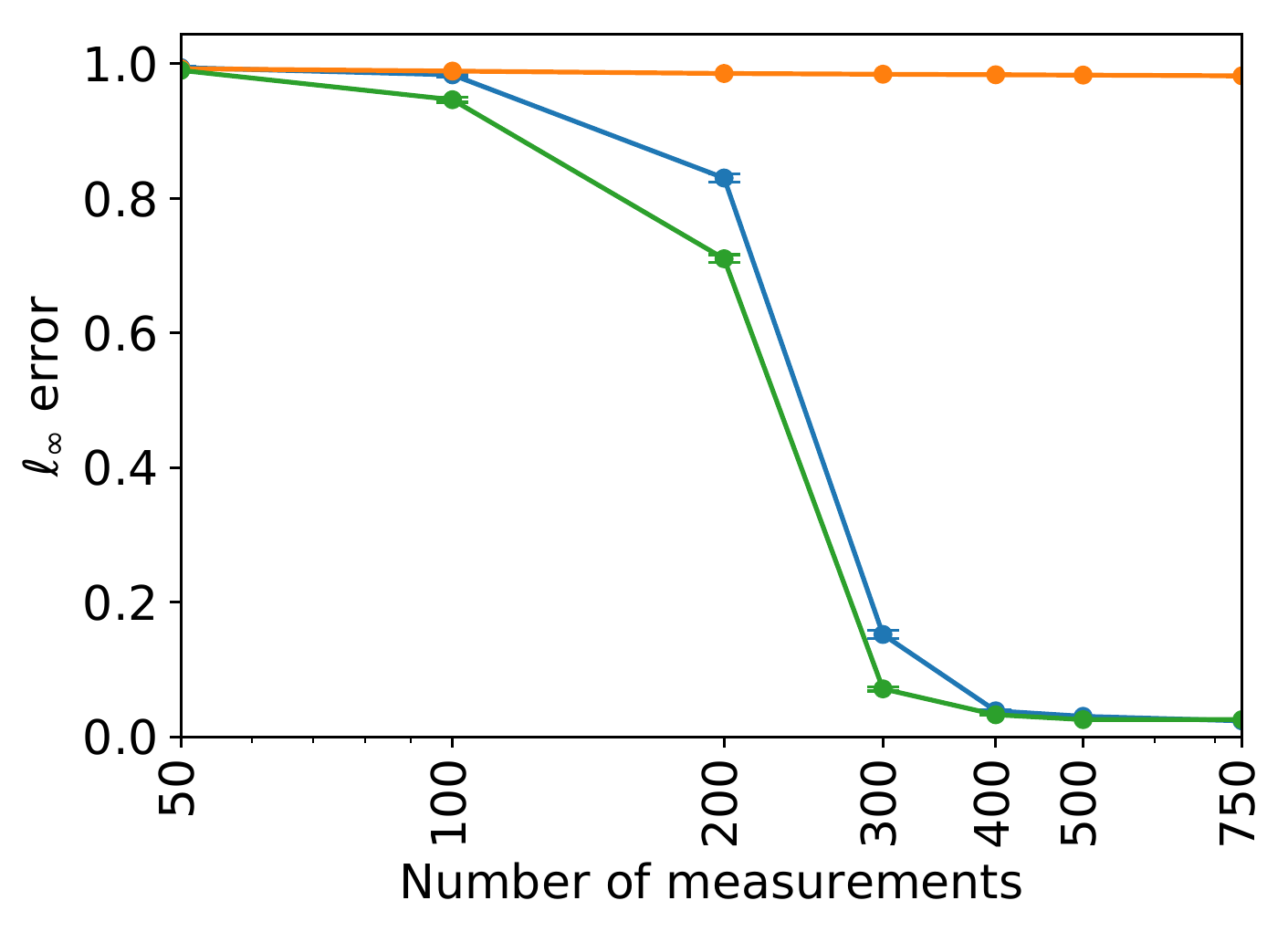}
\caption{Omniglot - $\ell_\infty$}
\end{subfigure}
\caption{Reconstruction error using $\ell_1$ (\textbf{left}), $\ell_2$ (\textbf{center}), and $\ell_\infty$ (\textbf{right}) norms for MNIST (\textbf{top}) and Omniglot (\textbf{bottom}). The performance of Sparse-Gen is better or competitive with both generative model-based methods and sparse vector recovery methods.}\label{fig:reconst}
\end{figure*} 

Our theoretical analysis formalizes the key properties of recovering signals using Sparse-Gen. As shown in Lemma~\ref{thm:decoder}, there exists a decoder for recovery based on such modeling assumptions that extends recovery guarantees based on vanilla sparse vector recovery and generative model-based recovery. 
Such recovery requires measurement matrices that satisfy both the RIP and S-REC conditions over the set of vectors that deviate in sparse directions from the range of a generative model function. In Theorems~\ref{thm:decoder_gaussian}-\ref{thm:decoder_gaussian_relu}, we observed that the number of measurements required to guarantee recovery with high probability grow almost linearly (with some logarithmic terms) with the latent space dimensionality $k$ of the generative model and the permissible sparsity $l$ for deviating from the range of the generative model. 
\section{Experimental Evaluation}

We evaluated Sparse-Gen for compressed sensing of high-dimensional signals from the domain of benchmark image datasets. Specifically, we considered the MNIST dataset of handwritten digits~\cite{lecun2010mnist} and the OMNIGLOT dataset of handwritten characters~\cite{lake2015human}. Both these datasets have the same data dimensionality ($28 \times 28$), but significantly different characteristics. The MNIST dataset has fewer classes (10 digits from 0-9) as opposed to Omniglot which shows greater diversity (1623 characters across 50 alphabets). Additional experiments with generative adversarial networks on the CelebA dataset are reported in the Appendix.

\textbf{Baselines.} We considered methods based on sparse vector recovery using LASSO~\cite{tibshirani1996regression,candes2005decoding} and generative model based recovery using variational autoencoders (VAE)~\cite{kingma-iclr2014,bora2017compressed}. 
For VAE training, we used the standard train/held-out splits of both datasets.
Compressed sensing experiments that we report were performed on the entire test set of images. 
The architecture and other hyperparameter details are given in the Appendix.
\begin{figure*}[ht]
\centering
\begin{subfigure}[b]{0.32\textwidth}
\centering
\includegraphics[width=\columnwidth]{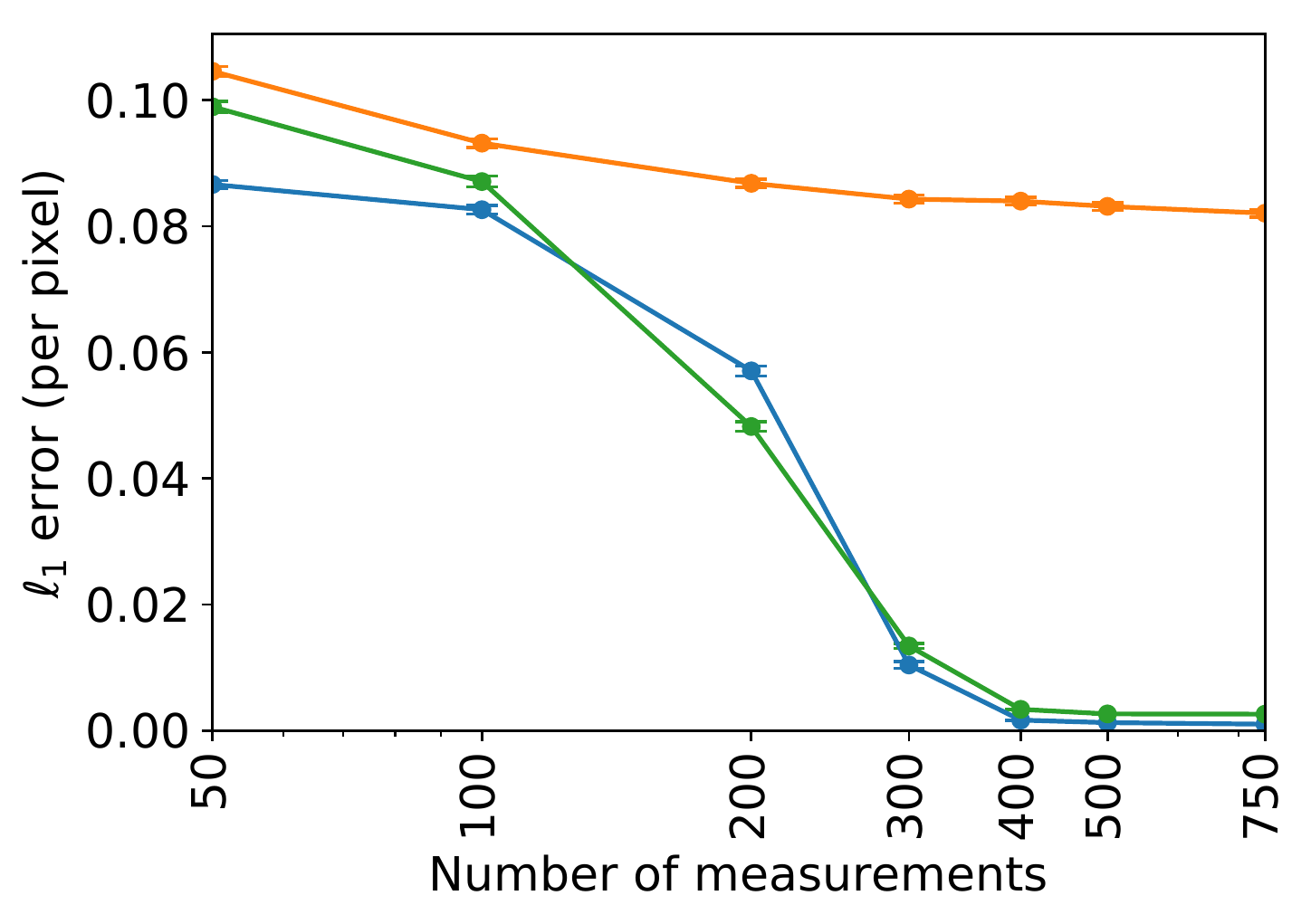}
\caption{mnist2omni - $\ell_1$}
\end{subfigure}
\begin{subfigure}[b]{0.32\textwidth}
\centering
\includegraphics[width=\columnwidth]{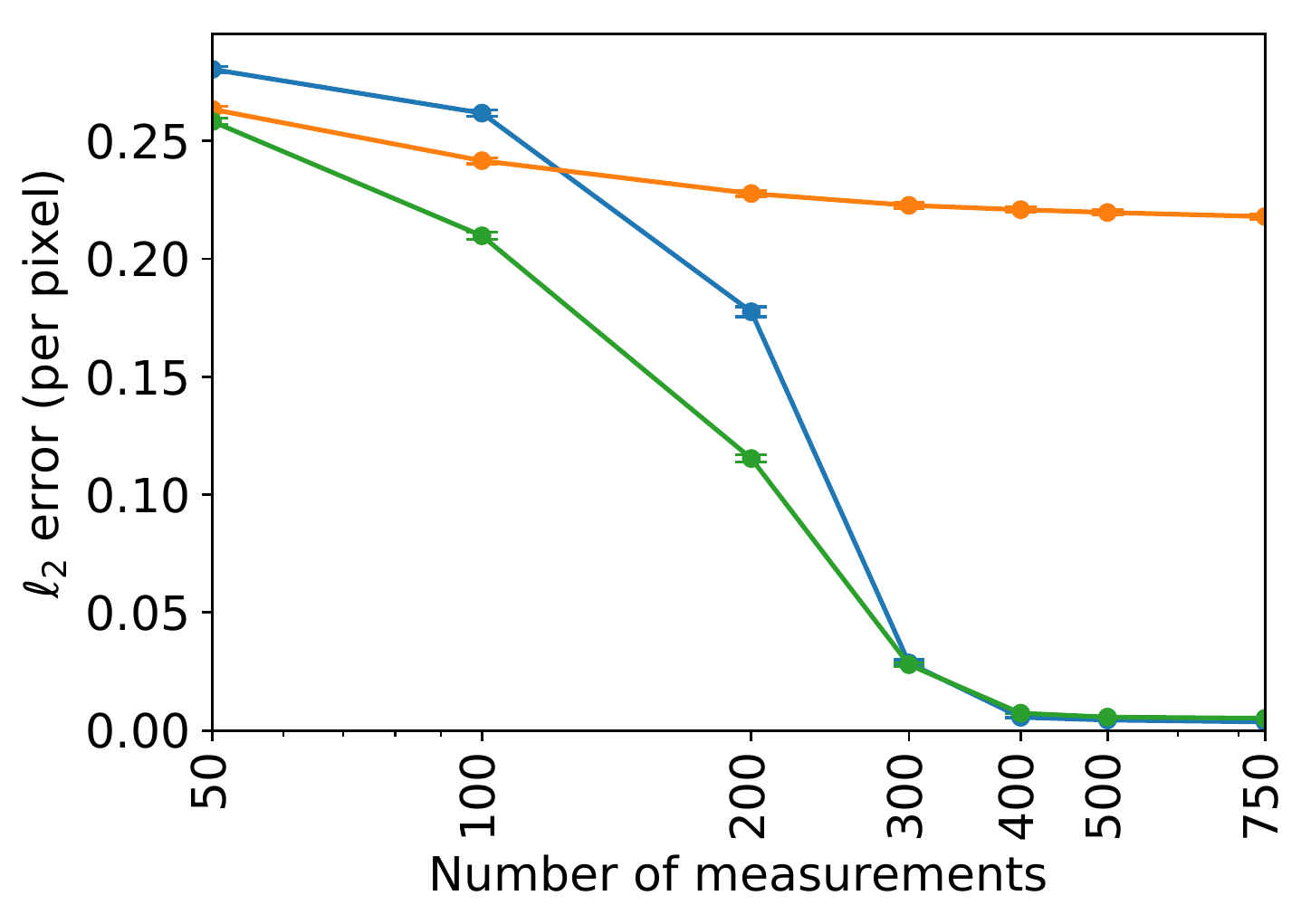}
\caption{mnist2omni - $\ell_2$}
\end{subfigure}
\begin{subfigure}[b]{0.32\textwidth}
\centering
\includegraphics[width=\columnwidth]{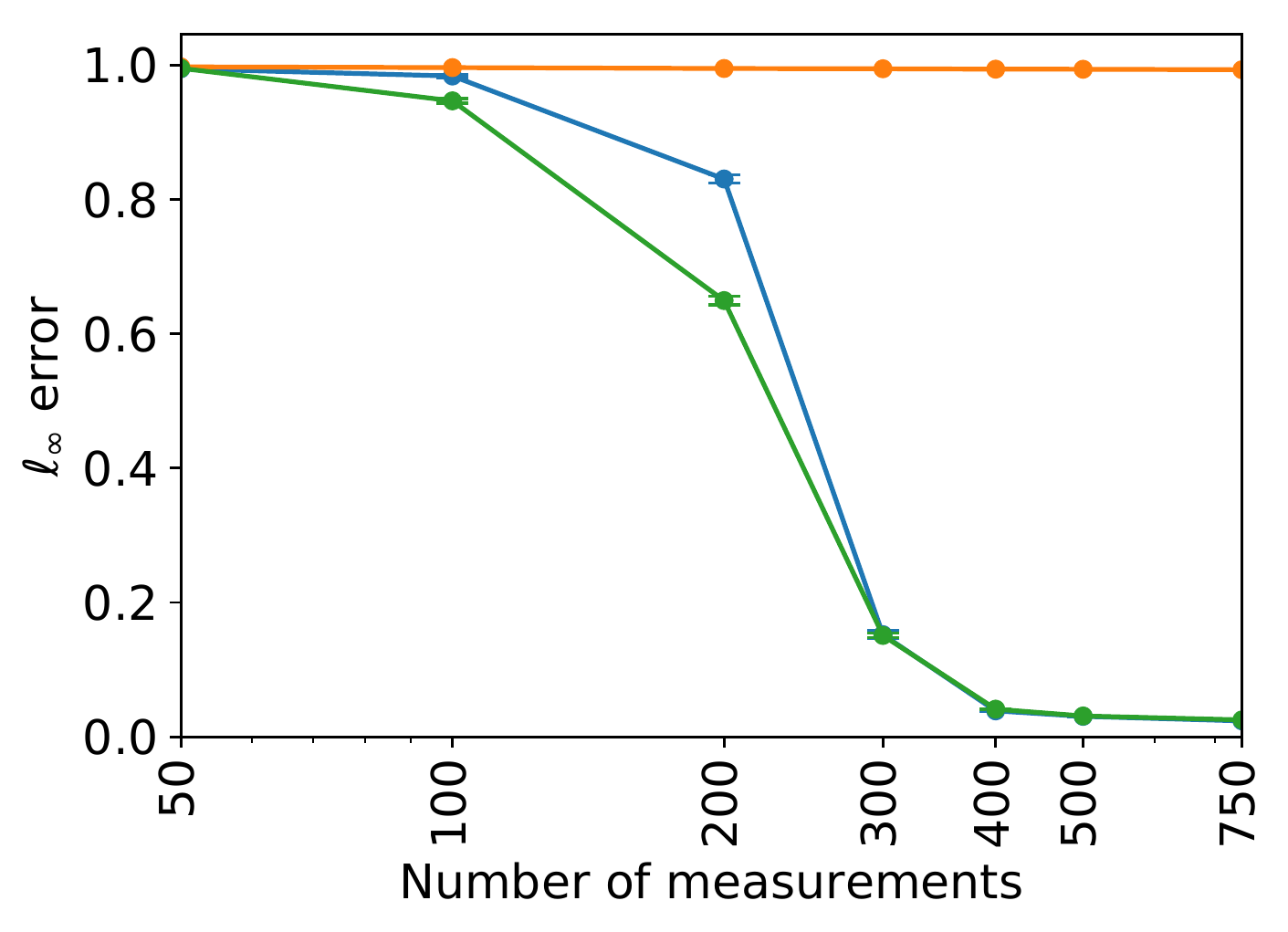}
\caption{mnist2omni - $\ell_\infty$}
\end{subfigure}
\begin{subfigure}[b]{0.32\textwidth}
\centering
\includegraphics[width=\columnwidth]{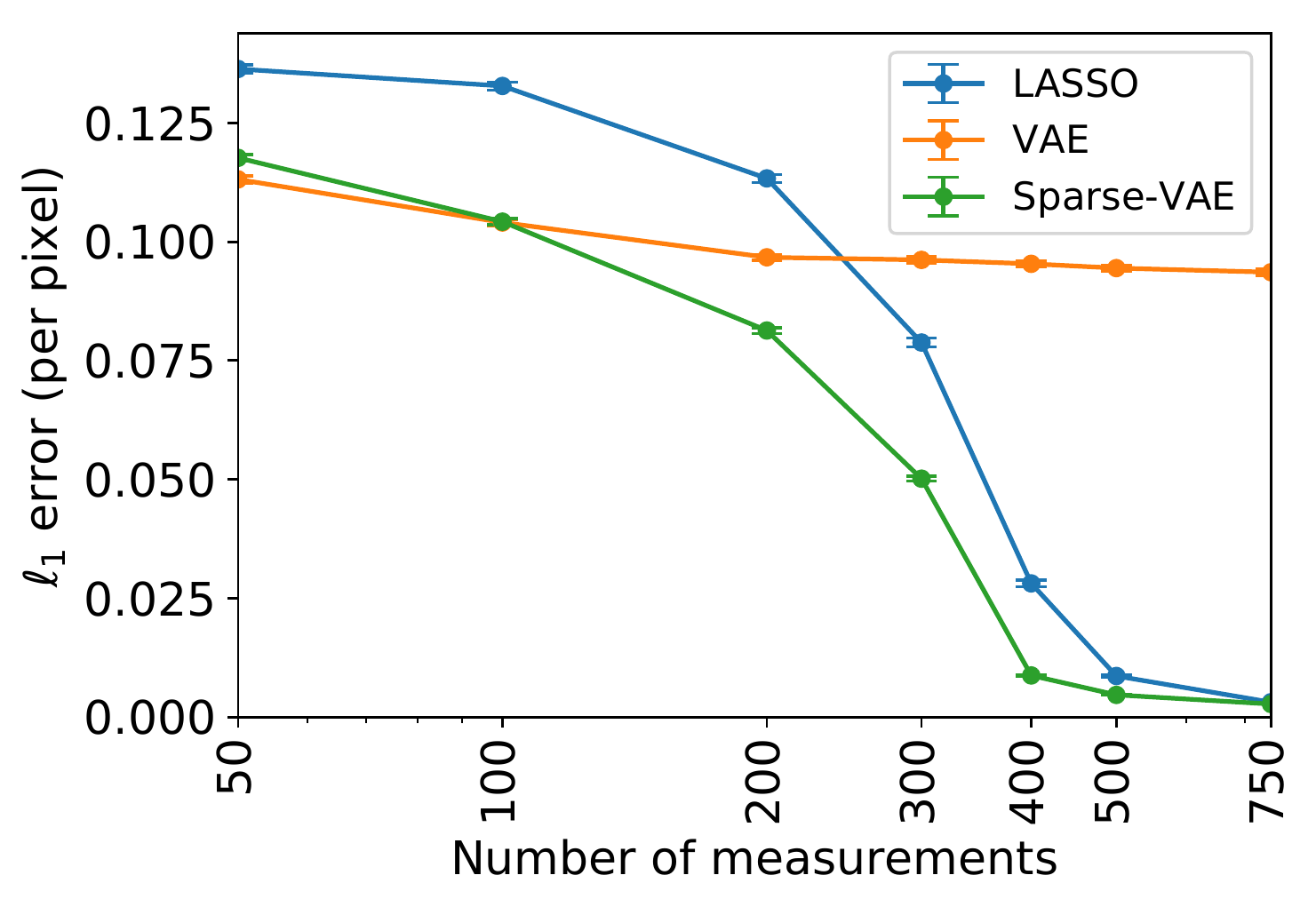}
\caption{omni2mnist - $\ell_1$}
\end{subfigure}
\begin{subfigure}[b]{0.32\textwidth}
\centering
\includegraphics[width=\columnwidth]{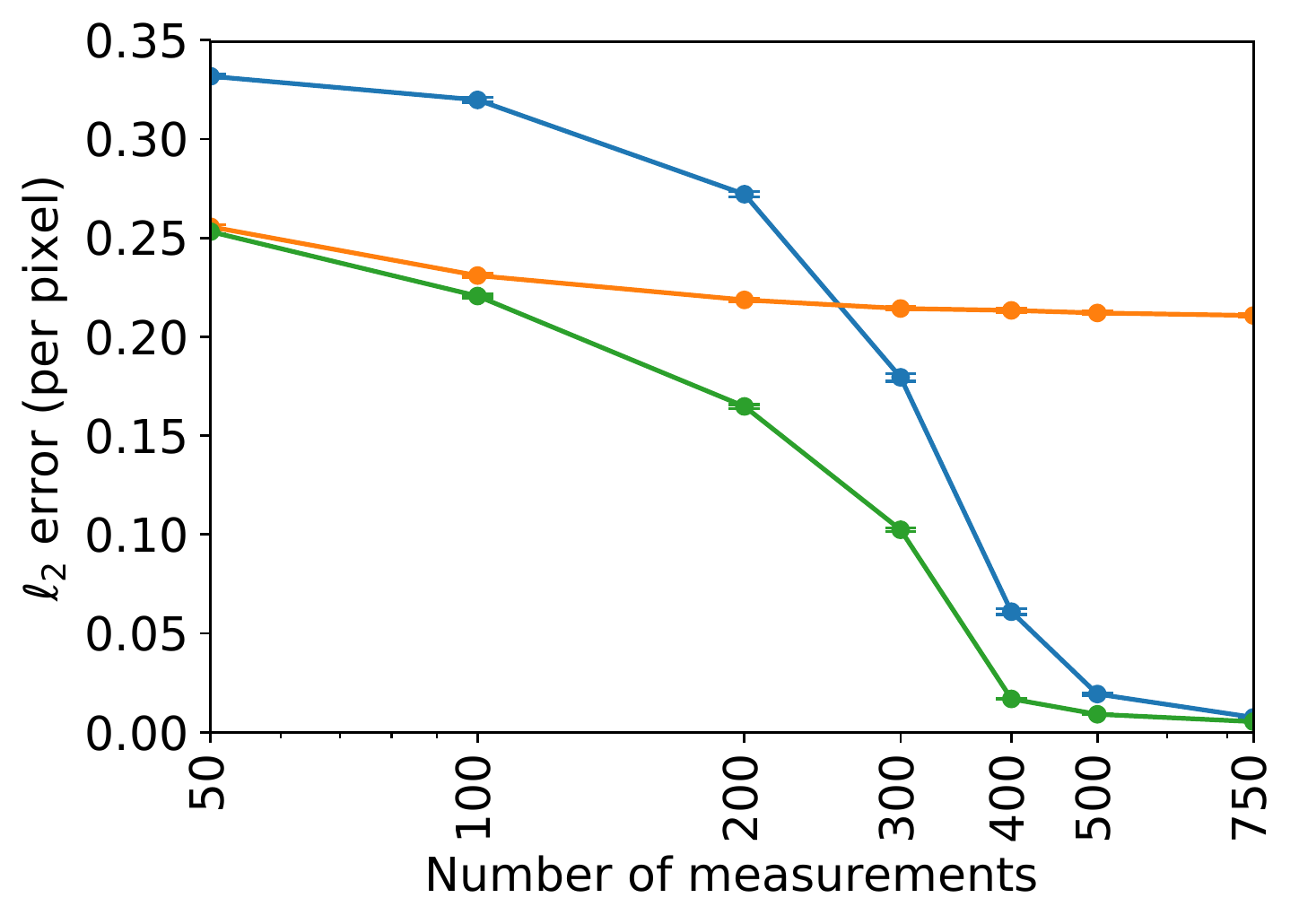}
\caption{omni2mnist - $\ell_2$}
\end{subfigure}
\begin{subfigure}[b]{0.32\textwidth}
\centering
\includegraphics[width=\columnwidth]{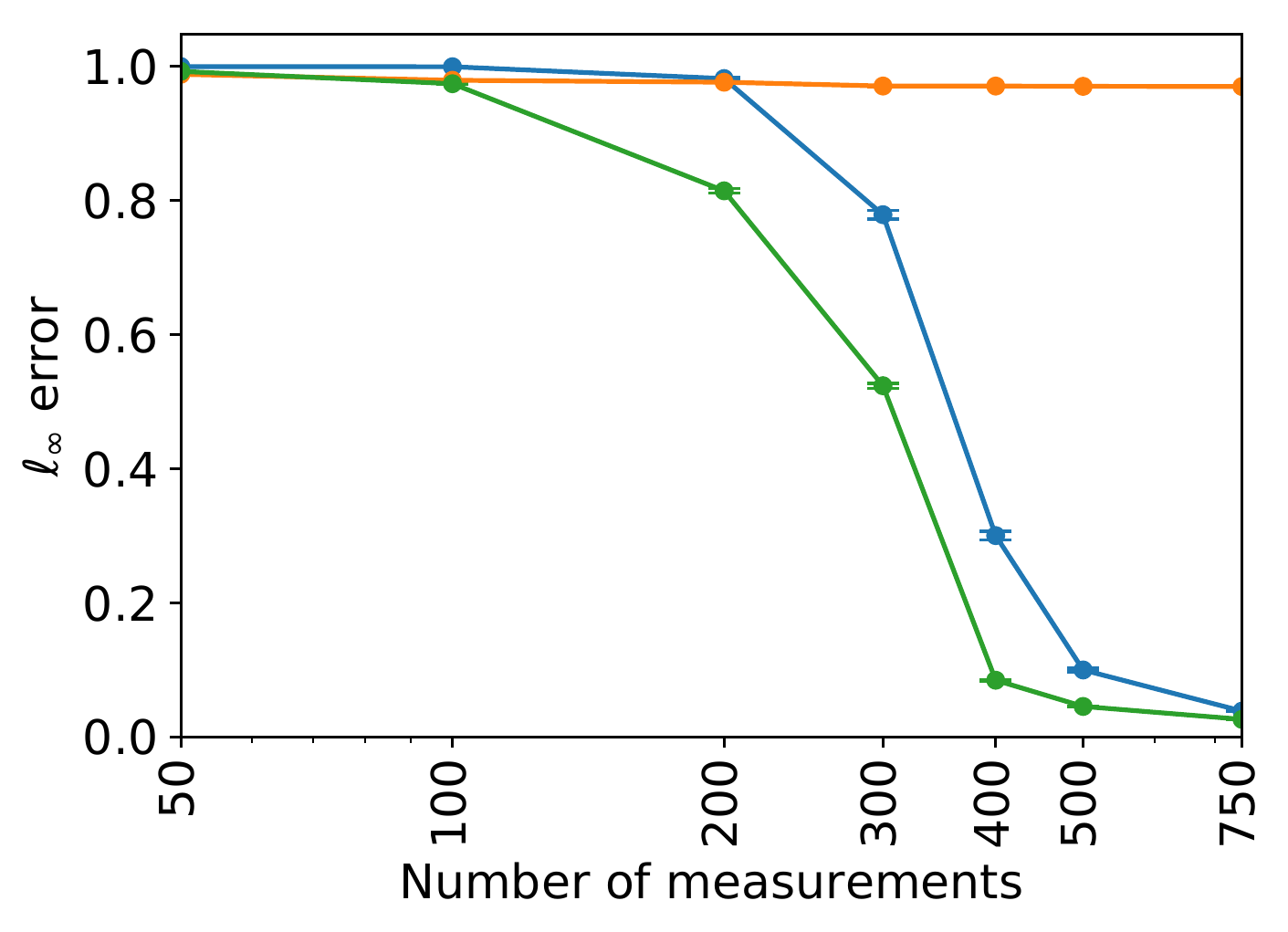}
\caption{omni2mnist - $\ell_\infty$}
\end{subfigure}
\caption{Reconstruction error in terms of $\ell_1$ (\textbf{left}), $\ell_2$ (\textbf{center}), and $\ell_\infty$ (\textbf{right}) norms for transfer compressed sensing from MNIST as source to Omniglot as target (\textbf{top}) and Omniglot as source and MNIST as target (\textbf{bottom}). Under the $\ell_2$ and $\ell_\infty$ metric, Sparse-VAE beats both VAE and LASSO. Under the $\ell_1$ metric, performance is slightly poorer until 100  measurements.}\label{fig:transfer}
\end{figure*}

\textbf{Experimental setup.} For the held-out set of instances, we artificially generated measurements $y$ through a random matrix $A \in \mathbb{R}^{m \times n}$ with entries sampled i.i.d. from a Gaussian with zero mean and standard deviation of $1/m$. Measurement noise is sampled from zero mean and diagonal scalar covariance matrix with entries as 0.01. For evaluation, we report the reconstruction error measured as $\Vert \widehat{x}- x \Vert_p$ where $\widehat{x}$ is the recovered signal and $p$ is a norm of interest, varying the number of measurements $m$ from $50$ to the highest value of $750$. We  report results for the $p=\{1,2, \infty\}$ norms.

We evaluated sensing of both continuous signals (MNIST) with pixel values in range $[0, 1]$ and discrete signals (Omniglot) with binary pixel values $\{0, 1\}$. For all algorithms considered, recovery was performed by optimizing over a continuous space. In the case of sparse recovery methods (including Sparse-Gen) it is possible that unconstrained optimization returns signals outside the domain of interest, in which case they are projected to the required domain by simple clipping, \textit{i.e.}, any signal less than zero is clipped to $0$ and similarly any signal greater than one is clipped to $1$.

\textbf{Results and Discussion.} The reconstruction errors for varying number of measurements are given in Figure~\ref{fig:reconst}. Consistent with the theory, the strong prior in generative model-based recovery methods outperforms the LASSO-based methods for sparse vector recovery. In the regime of low measurements, the performance of algorithms that can incorporate the generative model prior dominates over methods modeling sparsity using LASSO. The performance of plain generative model-based methods however saturates with increasing measurements, unlike Sparse-Gen and LASSO which continue to shrink the error. The trends are consistent for both MNIST and Omniglot, although we observe the relative magnitudes of errors in the case of Omniglot are much higher than that of MNIST. This is expected due to the increased diversity and variations of the structure of the signals being sensed in the case of Omniglot. We also observe the trends to be consistent across the various norms considered.

\begin{figure*}[ht]
\centering
\begin{subfigure}[b]{0.69\textwidth}
\centering
\includegraphics[width=\textwidth]{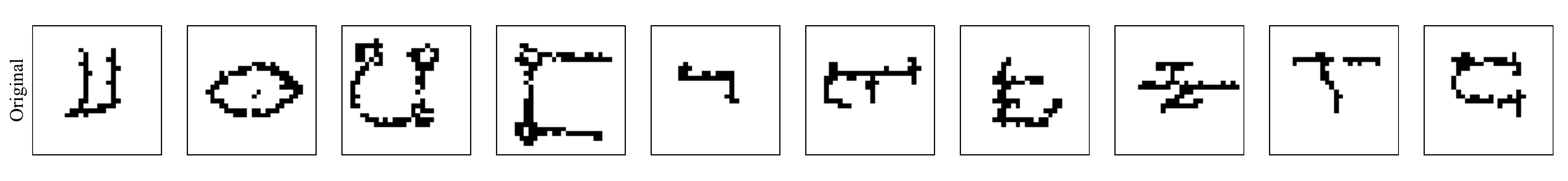}
\caption{Original}
\end{subfigure}
\begin{subfigure}[b]{0.69\textwidth}
\centering
\includegraphics[width=\textwidth]{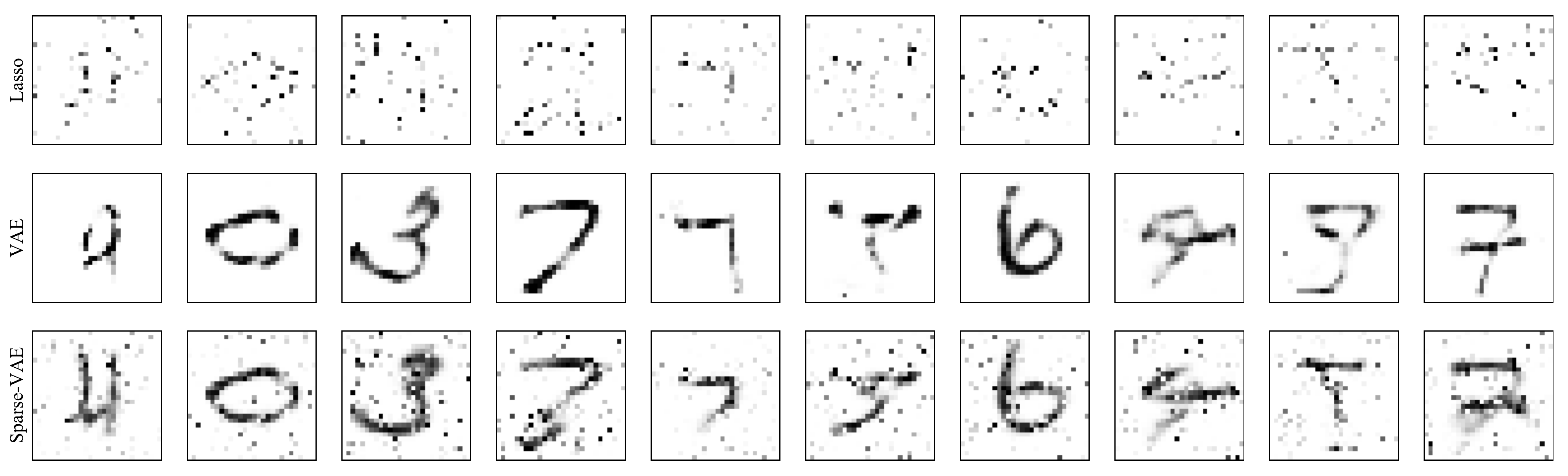}
\caption{Reconstructions with $m=100$ measurements.}
\end{subfigure}
\begin{subfigure}[b]{0.69\textwidth}
\centering
\includegraphics[width=\textwidth]{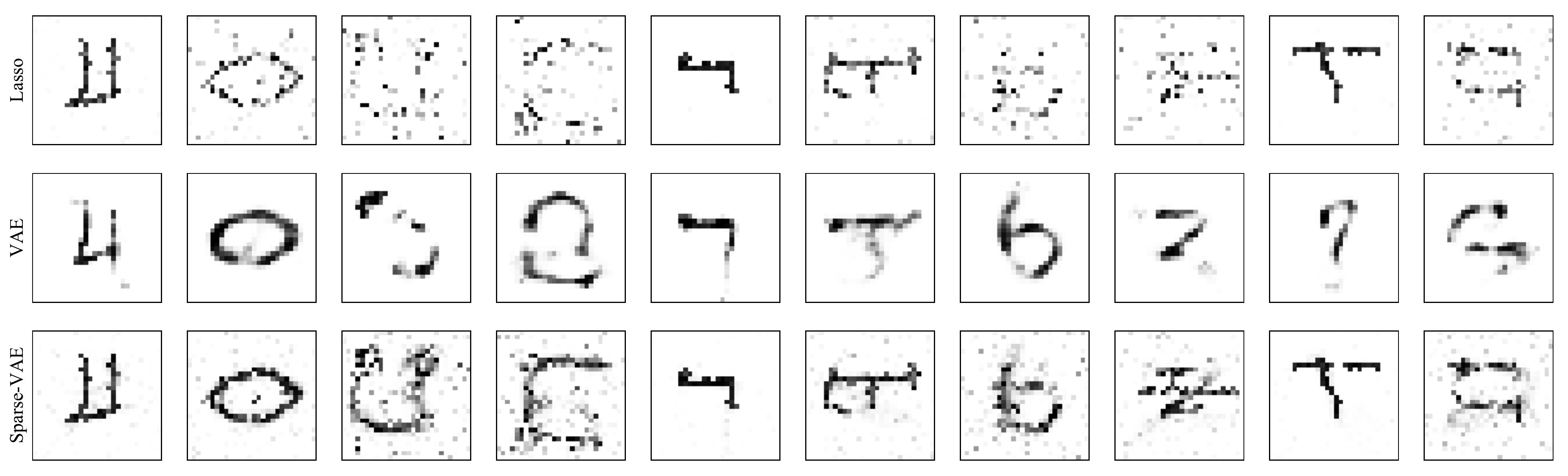}
\caption{Reconstructions with $m=200$ measurements.}
\end{subfigure}
\caption{The image matrix above shows the recovery of Omniglot images (\textbf{top}) using a VAE trained over MNIST using $100$ measurements (\textbf{middle}) and $200$ measurements (\textbf{bottom}). We note that with 100 measurements recovery is very poor across the board. With 200 measurements, all three methods show significant improvement and Sparse-VAE shows greater sensitivity to the domain being sensed.}\label{fig:transfer_q}
\end{figure*}

\subsection{Transfer compressed sensing}
One of the primary motivations for compressive sensing is to directly acquire the signals using few measurements. On the contrary, learning a deep generative model requires access to large amounts of training data. In several applications, getting the data for training a generative model might not be feasible. Hence, we test the generative model-based recovery on the novel task of \textit{transfer} compressed sensing.

\textbf{Experimental setup.} We train the generative model on a source domain (assumed to be data-rich) and related to a data-hungry target domain we wish to sense. Given the matching dimensions of MNIST and Omniglot, we conduct experiments transferring from MNIST (source) to Omniglot (target) and vice versa.

\textbf{Results and Discussion.} The reconstruction errors for the norms considered are given in Figure~\ref{fig:transfer}. For both the source-target pairs, we observe that the Sparse-Gen consistently performs well. Vanilla generative model-based recovery shows hardly an improvements with increasing measurements. We can qualitatively see this phenomena for transferring from MNIST (source) to Omniglot (target) in Figure~\ref{fig:transfer_q}. With only $m=100$ measurements, all models perform poorly and generative model based methods particularly continue to sense images similar to MNIST. On the other hand, there is a noticeable transition at $m=200$ measurements for Sparse-VAE where it adapts better to the domain being sensed than plain generative model-based recovery and achieves lower reconstruction error.

\section{Related Work}

Since the introduction of compressed sensing over a decade ago, there has been a vast body of research studying various extensions and applications~\cite{candes2005decoding,donoho2006compressed,candes2006robust}.  
This work explores the effect of modeling different structural assumptions on signals in theory and practice.

Themes around sparsity in a well-chosen basis has driven much of the research in this direction. For instance, the paradigm of model-based compressed sensing accounts for the interdependencies between the dimensions of a sparse data signal
~\cite{baraniuk2010model,duarte2011structured,gilbert2017towards}. Alternatively, adaptive selection of basis vectors from a dictionary that best capture the structure of the particular signal being sensed has also been explored~\cite{peyre2010best,tang2013compressed}. Many of these methods have been extended to recovery of structured tensors~\citep{zhang2013simultaneous,zhang2014hybrid}. In another prominent line of research involving Bayesian compressed sensing, the sparseness assumption is formalized by placing sparseness-promoting priors on the signals~\cite{ji2008bayesian,he2009exploiting,babacan2010bayesian,baron2010bayesian}.

Research exploring structure beyond sparsity is relatively scarce. Early works in this direction can be traced to ~\citet{baraniuk2009random} who proposed algorithms for recovering signals lying on a smooth manifold. The generative model-based recovery methods consider functions that do not necessarily define manifolds since the range of a generator function could intersect with itself. 
\citet{yu2011statistical} coined the term statistical compressed sensing and proposed algorithms for efficient sensing of signals from a mixture of Gaussians. The recent work in deep generative model-based recovery differs in key theoretical aspects as well in the use of a more expressive family of models based on neural networks. A related recent work by~\citet{hand2017global} provides theoretical guarantees on the solution recovered for solving non-convex linear inverse problems with deep generative priors. Empirical advances based on well-designed deep neural network architectures that sacrifice many of the theoretical guarantees have been proposed for applications such as MRI~\citep{mardani2017deep,mardani2017recurrent}. Many recent methods  propose to learn mappings of signals to measurements using neural networks, instead of restricting them to be linear, random matrices~\citep{mousavi2015deep,kulkarni2016reconnet,chang2017one,lu2018convcsnet}.  

Our proposed framework bridges the gap between algorithms that model structure using sparsity and enjoy good theoretical properties with advances in deep generative models, in particular their use for compressed sensing.
\section{Conclusion and Future Work}
The use of deep generative models as priors for compressed sensing presents a new outlook on algorithms for inexpensive data acquisition. In this work, we showed that these priors can be used in conjunction with classical modeling assumptions based on sparsity. Our proposed framework, Sparse-Gen, generalizes both sparse vector recovery and recovery using generative models by allowing for sparse deviations from the range of a generative model function. The benefits of using such modeling assumptions are observed both theoretically and empirically. 

In the future, we would like to design algorithms that can better model the structure \textit{within} sparse deviations. Follow-up work in this direction can benefit  from the vast body of prior work in structured sparse vector recovery~\cite{duarte2011structured}. From a theoretical perspective, a better understanding of the non-convexity resulting from generative model-based recovery can lead to stronger guarantees and consequently better optimization algorithms for recovery. Finally,  it would be interesting to extend Sparse-Gen for compressed sensing of other data modalities such as graphs for applications in network tomography and reconstruction~\citep{xu2011compressive}. Real-world graph networks are typically sparse in the canonical basis and can be modeled effectively using deep generative models~\citep{grover2018graphite}, which is consistent with the modeling assumptions of the Sparse-Gen framework.

\subsection*{Acknowledgements}
We are thankful to Tri Dao, Jonathan Kuck, Daniel Levy, Aditi Raghunathan, and Yang Song for helpful comments on early drafts. This research was supported by Intel Corporation, TRI, a Hellman Faculty Fellowship, ONR, NSF (\#1651565, \#1522054, \#1733686 ) and FLI (\#2017-158687). AG is supported by a Microsoft Research PhD Fellowship.

\bibliography{refs}
\bibliographystyle{icml2018}
\onecolumn
\appendix
\section{Proofs of theoretical results}

\subsection{Lemma~\ref{thm:decoder}}

To account for measurement noise in our analysis, we define the $\epsilon$-tube set $T$ of a matrix $A$ as,
$$\text{T}_{A}(\epsilon)=\{w:\|Aw\|_2\le \epsilon \}.$$
Note that in the absence of noise, $\text{T}_{A}(0)$ corresponds to the nullspace of $A$. Next, we define a difference function, $G':\iR^k\times\iR^k\rightarrow \iR^n$ such that $G'(z_1,z_2)=G(z_1)-G(z_2)$. 
Consequently, we obtain a difference set $S_{l,G'}$ as the Minkowski sum of $S_l(0)$ (the space of $l$ sparse vectors) and range of $G'$, 
\begin{align*}
S_{l,G'} = \cup_{z_1, z_2} S_l(G'(z_1, z_2)).
\end{align*} 
This allows us to define $\sigma_{l,G'}(x)$ as,
$$\sigma_{S_{l,G'}}(x)=\inf_{\hat{x}\in S_{l,G'}}\|x-\hat{x}\|_1.$$
Now, in order to prove Lemma~\ref{thm:decoder}, we state and derive a couple of lemmas. The proofs of the next two Lemmas (3 and 4) are modeled along the theory developed in ~\citet{cohen2009compressed} for the sensing of $l$-sparse vectors. We extend it to the case of $S_{l,G}$. Lemma 3 encodes the idea that for sensing to be successful any two points in $S_{l,G}$ should not be very close when acted upon by the measurement map $A$. This can be equivalently stated as requiring that any point in the nullspace of $A$ should not be approximated very well by points in $S_{2l,G'}$. Because we are working with bounded noise we need these results on the tube $T_A(2\epsilon)$, instead of just the nullspace. A point of interest is that informally the next lemma provides a sufficient condition for a good decoder to exist and also provides a different set of similar necessary conditions for good decoding. 
\begin{lemma}\label{thm:mixed_norm}
Given a measurement matrix $A\in \iR^{m\times n}$, measurement noise  $\epsilon$ such that $\|\epsilon\|_2\le \epsilon_{\max}$, and a generative model function $G:\iR^k\rightarrow \iR^n$ we want a decoder $\Delta:\iR^m\rightarrow \iR^n$ which provides the following $(\ell_2,\ell_1)$-mixed norm approximation guarantee on the set of $l$-sparse vectors $S_l$,
$$\|x-\Delta(Ax+\epsilon)\|_2 \le C_0l^{-t}\sigma_{l,G}(x) +C_1\epsilon_{\max}+\delta$$
for some constants $C_0,\delta,t\ge 0$. 

The sufficient condition for such a decoder to exist is given by,
$$\|\eta\|_2\le \frac{C_0}{2}l^{-t}\sigma_{2l,G'}(\eta)+C_1\epsilon_{\max}+\delta,\forall \eta\in T_A(2\epsilon_{\max}).$$
We call this the $(\ell_2,\ell_1)$-mixed norm null space property. 

A necessary condition for the same follows,
$$\|\eta\|_2\le C_0l^{-t}\sigma_{2l,G'}(\eta)+2C_1\epsilon_{\max}+2\delta,\forall \eta\in  T_A(\epsilon_{\max}).$$ 
\begin{proof} To prove the sufficiency of the null space condition, we define a decoder $\Delta:\iR^m\rightarrow \iR^n$ as follows,
$$\Delta(y)=\argmin\limits_{x:\|Ax-y\|_2\le \epsilon_{\max}}\sigma_{l,G}(x).$$
We will prove this decoder satisfies the mixed norm guarantee given the $(\ell_2,\ell_1)$-mixed norm null space property. Using the definition of $\Delta$, we have,
\begin{align*}
\|A(x-\Delta(Ax+\epsilon))\|_2&\le\|Ax+\epsilon-A\Delta(Ax+\epsilon))\|_2+\epsilon_{\max}\le 2\epsilon_{\max}.
\end{align*}
This implies $x-\Delta(Ax+\epsilon)\in T_A(2\epsilon_{\max})$. Combining with the mixed norm guarantee, we have,
\begin{align*}
\|x-\Delta(Ax+\epsilon)\|_2-\delta-C_1\epsilon_{\max} &\le \frac{C_0}{2}l^{-t}\sigma_{2l,G'}(x-\Delta(Ax+\epsilon))\\
&\le \frac{C_0}{2}l^{-t}(\sigma_{l,G}(x)+\sigma_{l,G}(\Delta(Ax+\epsilon)))\\
&\le C_0l^{-t}\sigma_{l,G}(x).
\end{align*}
The second last step follows from the triangle inequality whereby $\sigma_{2l,G'}(x+y)\le \sigma_{l,G}(x)+\sigma_{l,G}(y)$ and the last step uses the fact that the decoder is the minimizer of $\sigma_{l,G}(x)$.

For the necessary condition, consider any decoder $\Delta$ which provides the needed guarantee. Consider $\eta\in T_A(\epsilon_{\max})$ and now pick $z_0,z_1\in \iR^k,\eta_0\in S_{2l} $ 
such that the following inequality is satisfied,
\begin{align}\label{eq:eps_guarantee}\|\eta- (G(z_0)-G(z_1)+\eta_0)\|_1-\varepsilon' \le \sigma_{2l,G'}(\eta),\end{align}
where $\varepsilon'>0$. We can find a $z_0,z_1,\eta_0$ for any arbitrarily small and positive $\varepsilon'$. This is the case because we have, 
$$
\sigma_{2l,G'}(\eta)=\inf_{\hat{\eta} \in S_{2l,G'}}\|\eta-\hat{\eta}\|_1=\inf_{\hat{z}_0,\hat{z}_1\in \iR^k,\hat{\eta}_0\in S_{2l}}\|\eta- (G(\hat{z}_0)-G(\hat{z}_1)+\hat{\eta}_0)\|_1,
$$
which we obtain by parameterizing $\hat{\eta} \in S_{2l,G'}$ as $\hat{\eta}=G(\hat{z}_0)-G(\hat{z}_1)+\hat{\eta}_0$ for $\hat{z}_0,\hat{z}_1\in \iR^k,\hat{\eta}_0\in S_{2l}$. We cannot necessarily find $z_0,z_1,\eta_0$ such that $\varepsilon'=0$ because $S_{2l,G'}$ may not be a closed set. For convenience, we let $G_0=G(z_0)$ and $G_1=G(z_1)$ which means $G'(z_0,z_1)=G(z_0)-G(z_1)=G_0-G_1$.
We can split $\eta_0$ as $\eta_0=\eta_1+\eta_2$ for some $\eta_1,\eta_2\in S_{l}$, and for convenience define $\eta_3=\eta-\eta_0-G_0+G_1$. Note, we can now rewrite ~\eqref{eq:eps_guarantee} as,
\begin{align}\label{eq:eps_mod_guarantee}
\|\eta_3\|_1\le \sigma_{2l,G'}(\eta)+\varepsilon'.
\end{align}
Since $G_0+\eta_1\in S_{l,G}$, we have  $\sigma_{l,G}(G_0+\eta_1)=0$. This simplifies the $(\ell_2,\ell_1)$-mixed norm guarantee of our decoder when applied to $G_0+\eta_1$,
\begin{align}\label{eq:dec_guarantee}
\|G_0+\eta_1-\Delta(A(G_0+\eta_1))\|_2\le \delta+C_1\epsilon_{\max}.
\end{align}

Plugging in all the above, we have:
\begin{align*}
    \|\eta\|_2 &= \|\eta_2+\eta_1+\eta_3+G_0-G_1\|_2 \\
    &\le \|\eta_1+G_0-\Delta(A(\eta_1+G_0))\|_2 + \|\eta_3+\eta_2-G_1+\Delta(A(\eta_1+G_0))\|_2 \\
    &\le \delta +C_1\epsilon_{\max} + \|-\eta_3-\eta_2+G_1-\Delta(A\eta +A(G_1-\eta_2-\eta_3))\|_2 &&\text{(from Eq.~\eqref{eq:dec_guarantee})} \\
    &\le 2\delta + 2C_1\epsilon_{\max} + C_0l^{-t}\sigma_{l,G}(G_1-\eta_2-\eta_3) && (\text{since }\eta\in T_A(\epsilon_{\max})\text{ and the }(\ell_2,\ell_1)\text{-guarantee}) \\ 
    &\le 2\delta+2C_1\epsilon_{\max}+C_0l^{-t}\|\eta_3\|_1 \\
    &= C_0l^{-t}\sigma_{2l,G'}(\eta)+2\delta+2C_1\epsilon_{\max}+C_0l^{-t}\varepsilon'. && \text{(from Eq.~\eqref{eq:eps_mod_guarantee})}
\end{align*}
As we can make $\varepsilon'$ arbitrarily small we can make it tend to $0$ providing us with the required result.
\end{proof}
\end{lemma}
The next lemma basically shows that if $A$ satisfies the S-REC and RIP conditions then we operate in the constraint regime required by the previous lemma.
\begin{lemma}\label{thm:rip_srec}
If the measurement matrix $A \in \iR^{m \times n}$ satisfies S-REC($S_{(a+b)l/2,G'},1-\alpha,\delta$) and RIP($bl,\alpha$) for integers $a,b,l>0$ and function $G:\iR^k\rightarrow \iR^m$,  then we have for any vector $\eta\in T_A(\epsilon)$,
$$\|\eta\|_{2}\le (bl)^{-1/2}(C_0+1)\sigma_{al,G'}(\eta)+C_1\epsilon+\delta'$$
where $C_0=(1-\alpha)^{-1}(1+\alpha),C_1=(1-\alpha)^{-1}$, $\delta'=\delta(1-\alpha)^{-1}$.
\begin{proof} For any choice of $\eta\in T_A(\epsilon)$ and $G(z_1),G(z_2)$, let $\nu\in S_{al}$ be the minimizer of $\|\eta-G(z_1)+G(z_2)-\nu\|_1$. We can find this $\nu$ because $S_{al}$ is closed, concretely we can construct this $\nu$ by taking a $n$ dimensional vector which has everything but the top $al$ magnitude components in $\eta-G(z_1)+G(z_2)$ zeroed out. As the choice of $G(z_1)$ and $G(z_2)$ is arbitrary, it suffices to prove the statement for $\|\eta-G(z_1)+G(z_2)-\nu\|_1$ (instead of $\sigma_{al,G'}(\eta)$). 

Given a set of indices $\mathcal{I}$ for a $n$ dimensional vector we use $\mathcal{I}^c$ to denote the set of indices not in $\mathcal{I}$. Now note that $\nu$ corresponds to the $al$ largest coordinates of $\eta'=\eta-G(z_1)+G(z_2)$. Let the indices corresponding to those coordinates be $\mathcal{T}_0$. We take $\mathcal{T}_1$ to be the indices of the next $bl$ (and not $al$) largest coordinates. Similarly, define $\mathcal{T}_2,\ldots,\mathcal{T}_s$ to be subsequent indices for the next $bl$ largest coordinates. The final set $\mathcal{T}_s$ can contain indices of less than $bl$ coordinates. Let $\mathcal{T}_0\cup \mathcal{T}_1=\mathcal{T}$. We will use $x_\mathcal{I}$ to denote the vector obtained by zeroing out values in $x$ for all indices in the set $\mathcal{I}^c$.  
We can write $\eta_{\mathcal{T}}+(G(z_1)-G(z_2))_{\mathcal{T}^c}$ as $\eta_{\mathcal{T}}-(G(z_1)-G(z_2))_\mathcal{T}+(G(z_1)-G(z_2))$ where $ \eta_\mathcal{T},(G(z_1)-G(z_2))_\mathcal{T}\in S_{(a+b)l}$. We can write $\eta_\mathcal{T}+(G(z_1)-G(z_2))_\mathcal{T}$ as $s_1-s_2$ where $s_1,s_2\in S_{(a+b)l/2}$. This allows us to write $\eta_{\mathcal{T}}+(G(z_1)-G(z_2))_{\mathcal{T}^c}$ as $G(z_1)+s_1-(G(z_2)+s_2)$ where $G(z_1)+s_1,G(z_2)+s_2\in S_{(a+b)l/2,G'}$. Now we use the fact that $A$ satisfies S-REC($S_{(a+b)l/2,G'}$) to get,
\begin{align}\label{eq:signal_bound}
\|\eta_{\mathcal{T}}+(G(z_1)-G(z_2))_{\mathcal{T}^c}\|_{2}&= \|G(z_1)+s_1-(G(z_2)+s_2)\|_{2}\nonumber\\
&\le (1-\alpha)^{-1}\|A(G(z_1)+s_1-(G(z_2)+s_2))\|_{2} + (1-\alpha)^{-1}\delta &&(\text{using  S-REC})\nonumber\\
&\le (1-\alpha)^{-1}\|A(\eta_{\mathcal{T}}+(G(z_1)-G(z_2))_{\mathcal{T}^c})\|_{2} + (1-\alpha)^{-1}\delta. 
\end{align}
We can write $\eta=\eta_\mathcal{T}+\eta_{\mathcal{T}_2}+...+\eta_{\mathcal{T}_s}$. As $\eta \in T_A(\epsilon)$ we can write $A\eta_\mathcal{T}=-A(\eta_{\mathcal{T}_2}+...+\eta_{\mathcal{T}_s})+\gamma$ where $\|\gamma\|_2\le \epsilon$.
Hence,
\begin{align}\label{eq:measurement_bound}
\|A(\eta_\mathcal{T}+(G(z_1)-G(z_2))_{\mathcal{T}^c})\|_{2}&=\|A((\eta-G(z_1)+G(z_2))_{\mathcal{T}_2}+..+(\eta-G(z_1)+G(z_2))_{\mathcal{T}_s})-\gamma\|_{2}\nonumber\\
&=\|A\eta'_{\mathcal{T}_2}+...+A\eta'_{\mathcal{T}_s}-\gamma\|_2\nonumber\\
&\le \sum\limits_{j=2}^s\|A\eta'_{\mathcal{T}_j}\|_{2}+\|\gamma\|_2\nonumber\\
&\le(1+\alpha)\sum\limits_{j=2}^s\|\eta'_{\mathcal{T}_j}\|_{2}+\epsilon. &&(\text{using RIP})
\end{align}
From Eq.~\eqref{eq:signal_bound} and Eq.~\eqref{eq:measurement_bound}, we get,
$$\|\eta_{\mathcal{T}}+(G(z_1)-G(z_2))_{T^c}\|_{2}-\delta'\le(1-\alpha)^{-1}(1+\alpha)\sum\limits_{j=2}^s\|\eta'_{T_j}\|_{2}+(1-\alpha)^{-1}\epsilon.$$
Adding $\|\eta'_{\mathcal{T}^c}\|_{2}$ on both sides and applying the triangle inequality, we get,
\begin{align}\label{eq:midway}
\|\eta\|_{2}&\le\|\eta_{\mathcal{T}}+(G(z_1)-G(z_2))_{T^c}\|_{2}+\|\eta'_{\mathcal{T}^c}\|_{2}\nonumber\\
&\le((1-\alpha)^{-1}(1+\alpha)+1)\sum\limits_{j=2}^s\|\eta'_{\mathcal{T}_j}\|_{2} +\delta'+C_1\epsilon.
\end{align}
For any  $i\ge 1$, $j_1\in \mathcal{T}_{i+1}$ ,and $j_2\in \mathcal{T}_i$ we have $|\eta'_{j_1}|\le|\eta'_{j_2}|$ which in turn implies that $|\eta'_{j_1}|\le (bl)^{-1}\|\eta'_{\mathcal{T}_i}\|_{1}$. Squaring and adding the inequalities for all such indices in $T_{i}$ and $T_{i+1}$, we get,
$$\|\eta'_{i+1}\|_{2}\le (bl)^{-1/2}\|\eta'_{i}\|_{1}.$$
Substituting the result we obtained above in Eq.~\eqref{eq:midway}, we get,
$$\|\eta\|_{2}-\delta'-C_1\epsilon\le (bl)^{-1/2}((1-\alpha)^{-1}(1+\alpha)+1)\sum\limits_{j=1}^s\|\eta'_{\mathcal{T}_j}\|_{1}=(bl)^{-1/2}(C_0+1)\|\eta'_{\mathcal{T}_0^c}\|_{1}$$
finishing the proof.
\end{proof}
\end{lemma}

Lemma~\ref{thm:decoder} follows directly from Lemma~\ref{thm:mixed_norm} and Lemma~\ref{thm:rip_srec} after substituting  $a=1$ and $b=2$.

\subsection{Lemma~\ref{thm:gaussian}}
Recall that random Gaussian matrices satisfy RIP and S-REC properties with high probability~\cite{candes2005decoding,bora2017compressed}. For completeness and notation, we restate these facts before proving Lemma~\ref{thm:gaussian}.

\begin{fact}
Let $A \in \iR^{m \times n}$ be a random Gaussian matrix with each entry sampled i.i.d. from $\mathcal{N}(0,1/m)$.  $\alpha\in (0,1)$. For $$m=O\left(\frac{l}{\alpha^2}\log(n/l)\right),$$ $A$ satisfies RIP$(l,\alpha)$ with probability at least $1-e^{-\Omega(\alpha^2m)}$.
\end{fact}

\begin{fact}
Let $A \in \iR^{m \times n}$ be a random Gaussian matrix with each entry sampled i.i.d. from $\mathcal{N}(0,1/m)$. Let $G:\iR^k\rightarrow \iR^n$ be an $L$-Lipschitz function and define $B^k(r)=\{z:\|z\|_2\le r\}$ to be the $\ell_2$ norm ball. For
$$m=O\left(\frac{k}{\alpha^2}\log\left(\frac{Lr}{\delta}\right)\right), $$
$A$ satisfies S-REC($G(B^k(r)), 1-\alpha,\delta$) with probability at least $1-e^{-\Omega(\alpha^2 m)}$.
\end{fact}
Note the proofs of the next two results basically involve small modifications in the proofs presented in \citet{bora2017compressed} at a few key places to extend them from the setting of the range of the generative model $G$ to the set $S_{l,G}$.
\begin{proof}
We will use the mathematical constructs of $\epsilon$-nets for proving the lemma. Let $M$ be a $\delta/L$-net for $B^k(r)$. Then there exists a net such that, 
$$\log(|M|)\le k\log\left(\frac{Lr}{\delta}\right).$$
As this net is $\delta/L$-cover for $B^k(r)$, we will have that $G(M)$ is a $\delta$-cover of $G(B^k(r))$. \\

For any two points $z_1,z_2\in B^k(r)$ we can find points $z'_1,z'_2\in M$ such that distance in $\ell_2$ norm between $G(z_1)$ and $G(z'_1)$ is less than $\delta$ (similarly for $G(z_2)$ and $G(z'_2)$). Now consider some set of indices $I$ of size $l$ and $\nu$ be an $l$-sparse vector with support $I$ (that is all elements outside the indices in $I$ are zero). Using the triangle inequality, we get,
\begin{align*}
\|G(z_1)-G(z_2)+\nu\|_2&\le \|G(z_1)-G(z'_1)\|_2+\|G(z'_1)-G(z'_2)+\nu\|_2+\|G(z'_2)-G(z_2)\|_2\\
&\le \|G(z'_1)-G(z'_2)+\nu\|_2+2\delta.
\end{align*}
Again using the triangle inequality, we have,
$$\|AG(z'_1)-AG(z'_2)+A\nu\|_2\le\|AG(z'_1)-AG(z_1)\|_2+\|AG(z_1)-AG(z_2)+A\nu\|_2+\|AG(z_2)-AG(z'_2)\|_2.$$

From Lemma 8.3 in \citet{bora2017compressed}, we have $\|AG(z'_1)-AG(z_2)\|_2=O(\delta)$, and  $\|AG(z_2)-AG(z'_2)\|_2=O(\delta)$ with probability $1-e^{-\Omega(m)}$. Applying this to the previous inequality gives us,
$$\|AG(z'_1)-AG(z'_2)+A\nu\|_2\le\|AG(z_1)-AG(z_2)+A\nu\|_2+O(\delta).$$
We note for fixed $z'_1,z'_2$ and $\nu$ varying over points with support $I$, $G(z'_1)-G(z'_2)+\nu$ lie in a subspace of size at most $l+1$ (\textit{i.e.,} the subspace generated by $G(z'_1)-G(z'_2)$ and the basis for the subspace with support $I$). Using the machinery of oblivious subspace embeddings, we get,
$$(1-\alpha)\|G(z'_1)-G(z'_2)+\nu\|_2\le \|AG(z'_1)-AG(z'_2)+A\nu\|_2$$
will hold with probability $1-e^{-\Omega(\alpha^2m)}$ when $m=O(l/\alpha^2)$. We take a union bound over all choices of $z'_1,z'_2$ and choices of $I$ (choosing $l$ indices from $n$). Let the number of choices be $N$. 
Using the simple bound $\binom{n}{l} \le \Big(\frac{ne}{l}\Big)^l$ we have:
$$\log(N)\le 2\log(|M|)+l\log\left(\frac{en}{l}\right)\le 2k\log\left(\frac{Lr}{\delta}\right)+l\log\left(\frac{en}{l}\right).$$
 Now we conclude when,
$$m=O\Big(\frac{1}{\alpha^2}\Big(k\log\left(\frac{Lr}{\delta}\right)+l\log\left(\frac{n}{l}\right)\Big)\Big), $$
the following holds with probability $1-e^{-\Omega(\alpha^2m)}$ for all $z_1,z_2\in B^k(r)$ and $\nu\in S_l$ (the set of $l$-sparse vectors),
$$(1-\alpha)\|G(z_1)-G(z_2)+\nu\|_2\le \|A(G(z_1)-G(z_2)+\nu)\|_2+O(\delta).$$
The $O(\delta)$ can be scaled so that we just have $\delta$ there and that would not affect the bound on $m$ in the form it is stated.
\end{proof}

Finally, we note that Theorem~\ref{thm:decoder_gaussian} follows directly from the statements of Lemma~\ref{thm:decoder} and Lemma~\ref{thm:gaussian}.

\subsection{Theorem~\ref{thm:decoder_gaussian_relu}}
We first restate the full statement of Theorem~\ref{thm:decoder_gaussian_relu} for completeness:
\begingroup
\addtocounter{theorem}{-1}
\begin{theorem}
\textbf{(restated)} Let $G:\iR^k\rightarrow \iR^n$ be a neural network of depth $d$. For any 
$\alpha \in (0,1)$, $l>0$, let $A \in \R^{m\times n}$ be a random Gaussian matrix with 
\[
m=O\Big(\frac{1}{\alpha^2}\Big((k+l) d \log c+(k+l)\log(n/l)\Big)\Big).
\]
rows of i.i.d. entries scaled such that $A_{i,j} \sim N(0, 1/m)$.
Let $\Delta$ be the decoder satisfying Lemma~\ref{thm:decoder}. Then, we have with $1 - e^{-\Omega(\alpha^2 m)}$ probability,
$$\|x-\Delta(Ax+\epsilon)\|_2\le (2l)^{-1/2}C_0\sigma_{l,G}(x)+C_1 \epsilon_{\max}+\delta'$$
for all $x\in \R^n, \|\epsilon\|_2\le \epsilon_{\max}$, where $C_0=2((1+\alpha)(1-\alpha)^{-1}+1),C_1=2(1-\alpha)^{-1}$, and $\delta'=\delta(1-\alpha)^{-1}$.
\end{theorem}
\endgroup
The proof technique for Corollary~\ref{thm:decoder_gaussian_relu} is closely related to~\citet{matousek2002geom} and ~\citet{bora2017compressed}. We provide a geometrical proof sketch and refer the reader to the above works for further details.
\begin{proof}
Each individual layer of a neural network function $G$ consists of at most $c$ hyperplanes and the ReLU unit gets activated whenever the input of the previous layer crosses these hyperplanes. This implies that the partitions made by the hyperplanes on the input space of the previous layer describe regions where the function is defined by single matrix. From Lemma 8.3 of \citet{bora2017compressed}, the number of such partitions is at most $O(c^{k})$. Hence, the total number of partitions from the output space to the input space across $d$-layers will be $O(c^{kd})$. Consequently, the range of $G$ will be a union of $O(c^{kd})$ possibly truncated faces of dimension $k$ in $\iR^n$. 

Now if we consider the Minkowski sum $S_{l,G}$, then we observe that this set will be a union of $O(c^{kd}(n/l)^l)$ possibly truncated face of dimension $k+l$. Consider any two faces in $S_{l,G}$. The space defined by the difference of vectors (one from each face) will be part of a subspace of size $2k+2l+1$. This is because each face can be parameterized as $v_0+\sum t_ib_i$ where $v_0$ is fixed, $b_i$ is a basis for this face, and $t_i$ are the parameters. Hence the difference of two faces will have the same parametrization with at most $2k+2l$ basis vectors and a fixed point. Adding the fixed point to the basis gives us the required subspace. 

Finally, we use oblivious subspace embeddings to note that a random Gaussian matrix $A$ with each entry sampled from $\mathcal{N}(0,1/m)$ leads to a subspace emebedding with distortion $\alpha$ with a probability of $1-e^{-\Omega(\alpha^2m)}$ for $m=O((k+l)/\alpha^2)$. Since there are $O(c^{kd}(n/l)^l)$ such faces we take a union bound over all pairs of them to see that $A$ satisfies S-REC($S_{l,G},(1-\alpha)^{-1},0$) with probability $1-c^{2kd}(n/l)^{2l}e^{-\Omega(\alpha^2 m)}$. This implies that if we have,
$$m=O\left(\frac{1}{\alpha^2}\Big((k+l)(d\log c+\log(n/l))\Big)\right)$$
then $A$ satisfies S-REC($S_{l,G},(1-\alpha)^{-1},0$) with probability $1-e^{-\Omega(\alpha^2 m)}$ finishing the proof.
\end{proof} 

\section{Architectures and hyperparameter details}
For both the MNIST and Omniglot dataset, the network architecture was fixed to $784-500-500-20$ for both the generative network and the inference network (except for the final layer of the inference network which has $40$ units since both the mean and the variance of the Gaussian variational posterior are learned). Learning is done using Adam~\cite{kingma2014adam} with a learning rate of $0.001$.

For LASSO-based recovery the signal recovery algorithms/libraries were from CVXOPT~\cite{andersen2013cvxopt}. The Adam~\cite{kingma2014adam} implementations in Tensorflow~\cite{abadi2016tensorflow} was used for generative model-based recovery and Sparse-Gen recovery. The step sizes were selected by evaluating different step sizes via grid search over a held-out validation set (distinct from the held-out test set for which the scores are reported). The recovery procedure was run $10$ times each for the generative model based method and the Sparse-gen method, the value with the smallest measurement error is then returned.

\section{Additional results for CelebA dataset}
\begin{figure*}[!htb]
\begin{subfigure}[b]{0.32\textwidth}
\centering
\includegraphics[width=\columnwidth]{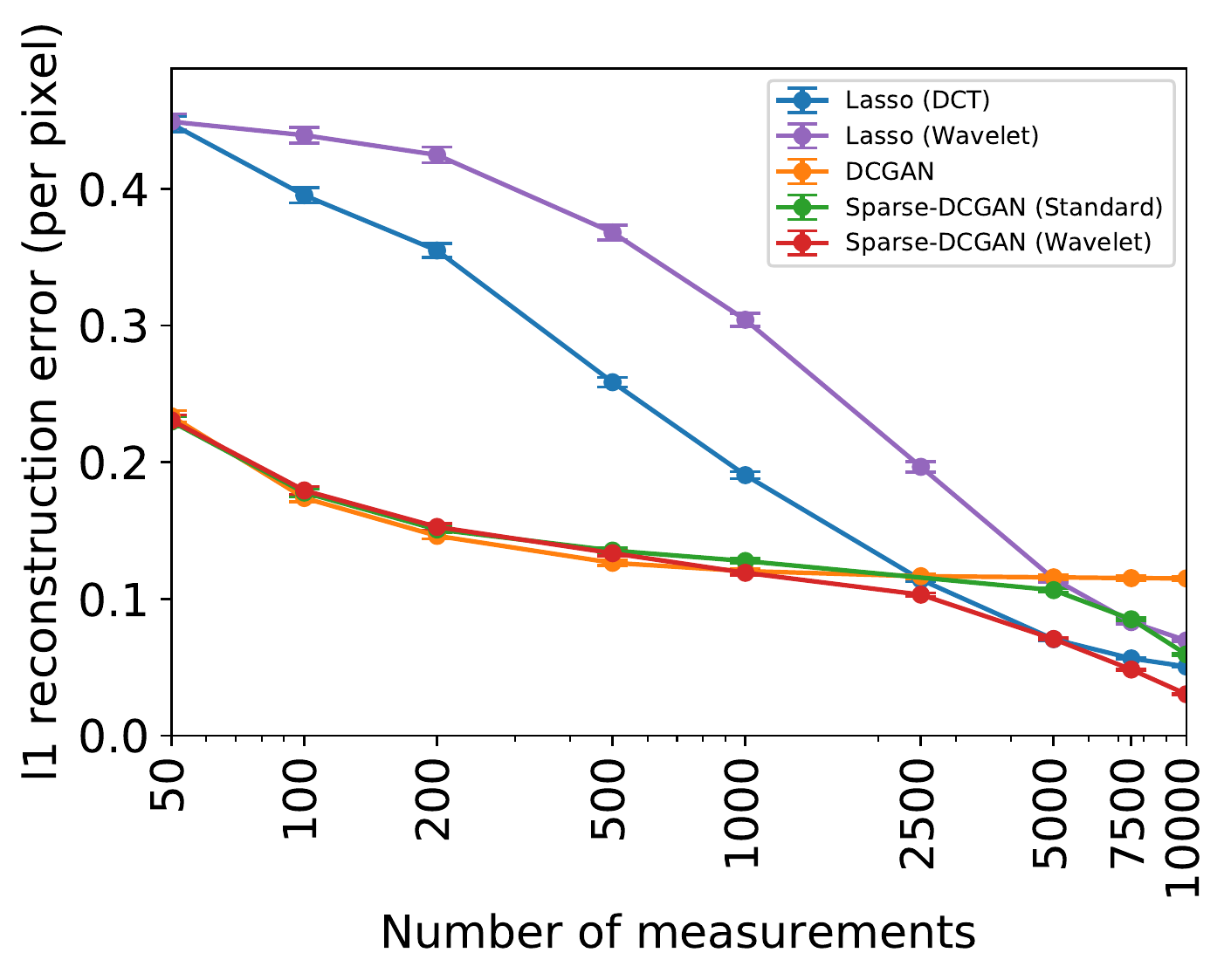}
\caption{CelebA - $\ell_1$}
\end{subfigure}
\begin{subfigure}[b]{0.32\textwidth}
\centering
\includegraphics[width=\columnwidth]{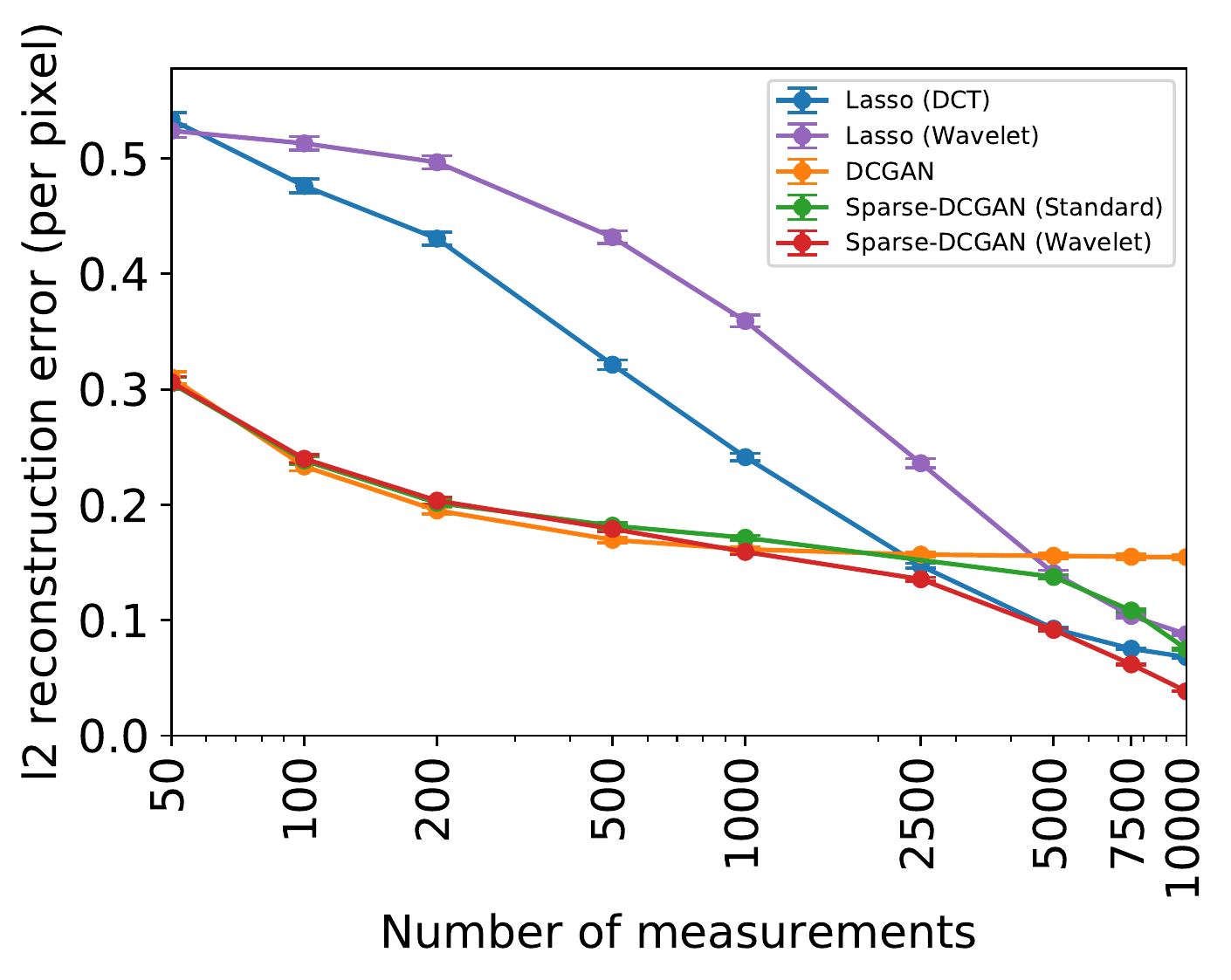}
\caption{CelebA - $\ell_2$}
\end{subfigure}
\begin{subfigure}[b]{0.32\textwidth}
\centering
\includegraphics[width=\columnwidth]{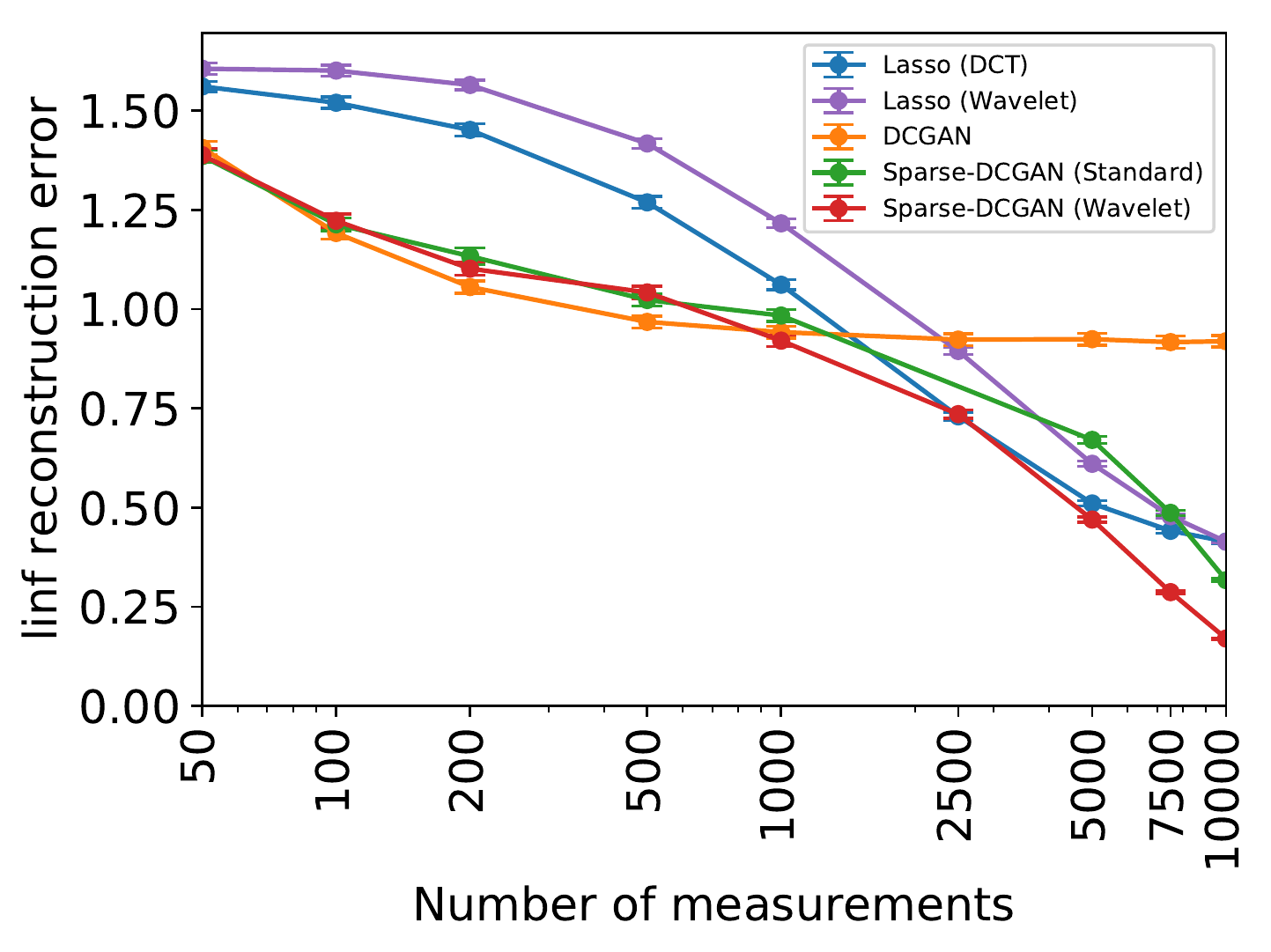}
\caption{CelebA - $\ell_\infty$}
\end{subfigure}
\caption{Reconstruction error in terms of $\ell_1$ (\textbf{left}), $\ell_2$ (\textbf{center}), and $\ell_\infty$ (\textbf{right}) norms for the CelebA datasets. The performance of Sparse-DCGAN is competitive with DCAN for low measurements and it matches Lasso at high measurements as expected.}\label{fig:reconst_celebA}
\end{figure*}
For the CelebA dataset, we train models based on the DCGAN \cite{radford2015unsupervised} architecture using adversarial training. 
As natural images are not sparse in the standard basis, we use basis vectors obtained from wavelets and discrete cosine transform for LASSO and Sparse-Gen (called Sparse-DCGAN here). The graphs show that the trends are similar to the MNIST and Omniglot experiments. Sparse-DCGAN shows comparable performance to DCGAN for low measurements and does better than LASSO and DCGAN as the number of measurements increase. The wavelet basis works better than the DCT basis for Sparse-DCGAN.

\end{document}